\documentclass[journal]{IEEEtran} 

\usepackage{amsmath,amsfonts,amssymb}
\usepackage{algorithmic}
\usepackage{algorithm}
\usepackage{array}
\usepackage[caption=false,font=normalsize,labelfont=sf,textfont=sf]{subfig}
\usepackage{textcomp}
\usepackage{stfloats}
\usepackage{url}
\usepackage{verbatim}
\usepackage{balance}
\usepackage{graphicx}
\graphicspath{{figures/}}
\usepackage{cite}

\usepackage{booktabs}

\usepackage{caption}
\usepackage{cprotect}

\usepackage{xcolor}
\usepackage{color}
\usepackage[colorlinks,urlcolor=blue,linkcolor=blue,citecolor=blue]{hyperref}

\usepackage{tikz}
\usetikzlibrary{shapes,arrows}
\usetikzlibrary{positioning,shadows,arrows}
\usetikzlibrary{calc}
\definecolor{mycolor}{rgb}{0.2,0.2,0.2}
\definecolor{mygreen}{rgb}{0,0.85,0}
\tikzstyle{input} = [ellipse,draw=black!15, fill=black!15,text width=2em,scale= 0.8, rotate=90, opacity=.4 ,text centered, rounded corners=1mm, minimum height=3.3em,minimum size=1em]
\tikzstyle{input2} = [ellipse,draw=green, fill=black!15,text width=2em,scale= 0.85, rotate=90, opacity=.4 ,text centered, rounded corners=1mm, minimum height=3.4em,minimum size=1.2em]
\tikzstyle{input5} = [ellipse,draw=green, fill=black!15,text width=2em,scale= 0.87, rotate=90, opacity=.4 ,text centered, rounded corners=1mm, minimum height=3.4em,minimum size=1.2em]
\tikzstyle{input4} = [ellipse,draw=green, fill=black!15,text width=2em,scale= 0.87, rotate=90, opacity=.4 ,text centered, rounded corners=1mm, minimum height=3.4em,minimum size=2.4em]
\tikzstyle{input3} = [draw,draw=black!30, fill=white,text width=16.4em,scale= 0.9, text centered, rounded corners=1mm, minimum height=.3em,minimum size=1.1em]
\tikzstyle{input6} = [draw,draw=black, densely dashed,fill=white,text width=17em,scale= 1, opacity=.2, rounded corners=1mm, text centered, minimum height=3em,minimum size=1.1em]
\tikzstyle{line1} = [draw,  color=green, -latex']
\tikzstyle{line2} = [draw,  color=black, -latex']
\tikzstyle{tri} = [draw=black!10, shape border rotate=30, regular polygon, regular polygon sides=3, fill=black!10, scale = .422,node distance=2.5cm, minimum height=2em]
\tikzstyle{tri1} = [draw=black!10, shape border rotate=30, regular polygon, regular polygon sides=3, fill=black!10, scale = .49,node distance=2.5cm, minimum height=2em]
\tikzstyle{trm} = [rectangle,draw=white, fill=white,text width=.3em, scale=.5, text centered, minimum height=1em]
\tikzstyle{crc} = [circle,draw=black!30, fill=white,text width=.3em, scale=.4, text centered, minimum height=1em]
\tikzstyle{trm1} = [rectangle,thin,draw=black!20, fill=white,text width=2.5em, scale=.5, text centered, minimum height=1em]
\tikzstyle{trm2} = [ellipse,draw=green!60, fill=black!10,text width=3.3em, scale=.45 ,text centered, rounded corners=1mm, minimum height=1em]
\tikzstyle{trm3} = [ellipse,draw=green!60, fill=black!10,text width=4em, scale=.45 ,text centered, rounded corners=1mm, minimum height=1em]
\tikzstyle{trm4} = [ellipse,draw=green!60, fill=black!10,text width=5.5em, scale=.41 ,text centered, rounded corners=1mm, minimum height=1em]
\tikzstyle{tra} = [rectangle,draw=white, fill=white,text width=6.1em, scale=.5, text centered, minimum height=1em]
\tikzstyle{tras} = [rectangle,draw=white, fill=white,text width=3em, scale=.5, text centered, minimum height=1em]
\tikzstyle{trat} = [rectangle,draw=white, fill=white,text width=5em, scale=.5,rotate = 90, minimum height=1em]

\tikzstyle{boxisb} = [draw,draw=red!60, fill=white,text width=6em,scale= 0.9, text centered, minimum height=.3em,minimum size=1.1em]

\usepackage{gensymb} 
\usepackage{float}
\usepackage{wrapfig}
\usepackage{bm}
\usepackage{tabularx}
  \newcolumntype{Y}{>{\centering\arraybackslash}X}

\usepackage{lineno}

\usepackage[normalem]{ulem} 

\newcommand{\be}{\begin{equation}}
\newcommand{\ee}{\end{equation}}
\newcommand{\ba}{\begin{array}}
\newcommand{\ea}{\end{array}}

\newcommand{\blockcomment}[1]{}

\begin{document}

\title{Centimeter-level Positioning by Instantaneous Lidar-aided GNSS Ambiguity Resolution}

\author{
    Junjie Zhang,
    Amir Khodabandeh
    and
    Kourosh Khoshelham\\
    Submitted to: {\it Meas. Sci. Technol.}
    \thanks{J.~Zhang, A.~Khodabandeh and K.~Khoshelham are with the Department~of~Infrastructure~Engineering, The~University~of~Melbourne, Parkville, VIC 3010, Australia (e-mail: \mbox{junjiez@student.unimelb.edu.au}; \mbox{akhodabandeh@unimelb.edu.au}; \mbox{k.khoshelham@unimelb.edu.au}).}
}

\maketitle

\begin{abstract}
High-precision vehicle positioning is key to the implementation of modern driving systems in urban environments. Global Navigation Satellite System (GNSS) carrier phase measurements can provide millimeter- to centimeter-level positioning, provided that the integer ambiguities are correctly resolved. Abundant code measurements are often used to facilitate integer ambiguity resolution (IAR), however, they suffer from signal blockage and multipath in urban canyons. In this contribution, a lidar-aided instantaneous ambiguity resolution method is proposed. Lidar measurements, in the form of 3D keypoints, are generated by a learning-based point cloud registration method using a pre-built HD map and integrated with GNSS observations in a mixed measurement model to produce precise float solutions, which in turn increase the ambiguity success rate. Closed-form expressions of the ambiguity variance matrix and the associated Ambiguity Dilution of Precision (ADOP) are developed to provide a priori evaluation of such lidar-aided ambiguity resolution performance. Both analytical and experimental results show that the proposed method enables successful instantaneous IAR with limited GNSS satellites and frequencies, leading to centimeter-level vehicle positioning.
\end{abstract}


\section{Introduction \label{sec:intro}}

With
the growing interest in the development of autonomous driving systems, accurate positioning of vehicles in urban environments becomes the fundamental requirement to accomplish such applications. In order to achieve the current goal of autonomous driving, namely the \mbox{Level 4} systems which should enable driverless operations in a restricted domain, decimeter- to centimeter-level positioning accuracy is required to maintain a vehicle on the road within its lane~\cite{joubert2020developments}. It has been suggested that for typical road geometries and driving scenarios, the positioning accuracy of a passenger vehicle on urban roads should be at least 10 cm, 10 cm and 48 cm in the lateral (across the road), longitudinal (along the road) and vertical directions, respectively~\cite{reid2019localization}. Realizing and maintaining such high accuracy does therefore demand a rigorous integration of multiple measuring sensors. As a result, an array of multiple measuring sensors including, but not limited to, Global Navigation Satellite System (GNSS) receivers and Light Detection and Ranging (lidar) devices, is commonly found on modern vehicles. While GNSS is considered an essential positioning technology as it provides globally referenced solutions, its signals suffer from blockages and multipath effects in cities due to the high density of buildings~\cite{hofm08,liu2021distributed}. This is the more so as high-precision GNSS positioning requires the utilization of multi-frequency carrier phase signals transmitted by a modest number of visible satellites, a condition that is challenging to be met in urban areas~\cite{xiong2020carrier}. In contrast, lidar is not subject to the aforementioned error sources, providing abundant measurements in cities due to the existence of rich geometric features, yet it only offers locally referenced positioning solutions in standalone mode~\cite{zhang2021continuous}. In this contribution, we therefore aim to exploit the complimentary advantages of GNSS and lidar for urban positioning with a particular focus on the ultra-precise {\em carrier phase} measurements.

Although GNSS code (pseudo-range) measurements are easily accessible and served for standard positioning services, it is their carrier phase counterparts that can deliver precise parameter solutions~\cite{hofm08,blew,teunissen1995least,verhagen2005reliability,Gunther12,khodabandeh2021study}. GNSS carrier phase measurements are approximately two orders of magnitude more precise than the corresponding code versions, and are crucial for achieving centimeter-level positioning. The challenge of using carrier phase measurements, however, is that they are biased by a) unknown integer-valued ambiguities and b) instrumental phase delays~\cite{teunissen1995least}. The latter can be eliminated by the widely used positioning technique of real-time kinematic (RTK). In RTK, observations of a nearby reference GNSS station are subtracted from those of the to-be-positioned receiver to remove or largely reduce common sources of GNSS errors like clock offsets, instrumental biases and atmospheric delays~\cite{ALKAN2020107995}. However, the former, i.e. the unknown integer ambiguities, have to be estimated as real-valued float solutions first. Depending on the precision of the float solutions, Integer ambiguity resolution (IAR) is then employed to map such float ambiguities to their correct integers. If performed successfully, IAR yields ambiguity-resolved carrier phase measurements that can act like ultra-precise code measurements for positioning.

Whether or not IAR is successful is determined by the probability of correct integer estimation, the so-called ambiguity `success rate'~\cite{teunissen1999optimality}. The success rate is driven by the float ambiguity variance matrix, which on its turn, is governed by the number and precision of the measurements. In the event that the ambiguity success rate is low, one must refrain from fixing the float ambiguities as it often leads to unacceptably large positioning errors. On the other hand, for the vehicle positioning case where the location of the moving GNSS receiver is highly varying in time, successful IAR is required to be carried out in an {\em instantaneous} manner so as to maintain continuous centimeter-level positioning. Instantaneous or single-epoch IAR is only made possible when a large number of measurements from multiple satellite systems/frequencies are available~\cite{teunissen2017carrier}, which is often not the case in densely built-up areas due to signal blockage. In such environments, {\em complementary} sensing devices are needed for additional measurement.

As a prominent example of such complementary devices, lidar is capable of collecting feature-rich point clouds of the surrounding environments in urban canyons, providing 3D point maps which can be used for mitigation of GNSS errors, see, e.g.,~\cite{wen2019correcting,wen2020object,chiang2020performance}.
In particular, lidar's aiding role in GNSS IAR has been investigated by the recent studies~\cite{qian2020lidar,qian2020cooperative,li2020high,li2021feature}, which rely on features described by geometric characteristics that can be unavailable for `irregular' structures. Meanwhile, recent advances in deep learning have offered data-driven approaches for producing point feature descriptors which are more invariant and robust against geometric complexity than those from the traditional methods, thus improving the chance of successful point cloud registration~\cite{zhang2020deep,grigorescu2020survey}. The goal of the present contribution is to leverage such learning-based lidar features for high-precision positioning and present a GNSS-lidar integration method so as to realize successful instantaneous IAR.
To enable an a priori prediction of the method's IAR performance, we develop closed-form analytical expressions for both the lidar-aided float ambiguity variance matrix and the associated Ambiguity Dilution of Precision (ADOP)~\cite{teunissen1997canonical,odijk2008adop}. As an intrinsic measure for the average precision of the float ambiguities, ADOP indicates the extent to which the mapping of the float ambiguities to their integers is successful. The applicability of analytical IAR measures presented are supported by experimental results, showing how lidar measurements provide conditions under which instantaneous IAR becomes successful even for the single-frequency GNSS data of low-cost receivers.

The main contributions of this research include: 1) lidar measurements generated via a learning-based keypoint extraction strategy are integrated with GNSS code and carrier phase measurements at the observation-level in a mixed measurement model, enabling successful IAR for restricted GNSS satellite visibility; 2) closed-form ADOP expressions are developed to predict the performance of lidar-aided IAR. The remainder of this paper proceeds as follows. In Section \ref{sec:lidar}, a lidar-based method using a high-definition (HD) map via deep learning for aiding GNSS ambiguity resolution is proposed. Section~\ref{sec:gnss} presents the mixed measurement model with which one can integrate the stated lidar observations with their GNSS counterparts. Analytical measures for the precision of the lidar-aided float ambiguities are then presented in Section~\ref{sec:analysis}. We thereby show how the precision of the integrated solution competes with that of the lidar-only solution to reduce the ADOP, thus improving the corresponding IAR performance. Section~\ref{sec:setup} presents the configurations of the simulated experiment that is followed by Section~\ref{sec:results} showing the corresponding numerical results. A discussion on the results is given in Section~\ref{sec:discussion}. Finally, concluding remarks are drawn in Section~\ref{sec:concl}.

\section{Lidar observations generated by point cloud registration \label{sec:lidar}}

\begin{figure}[!t]
    \centering
    \includegraphics[width=.9\linewidth]{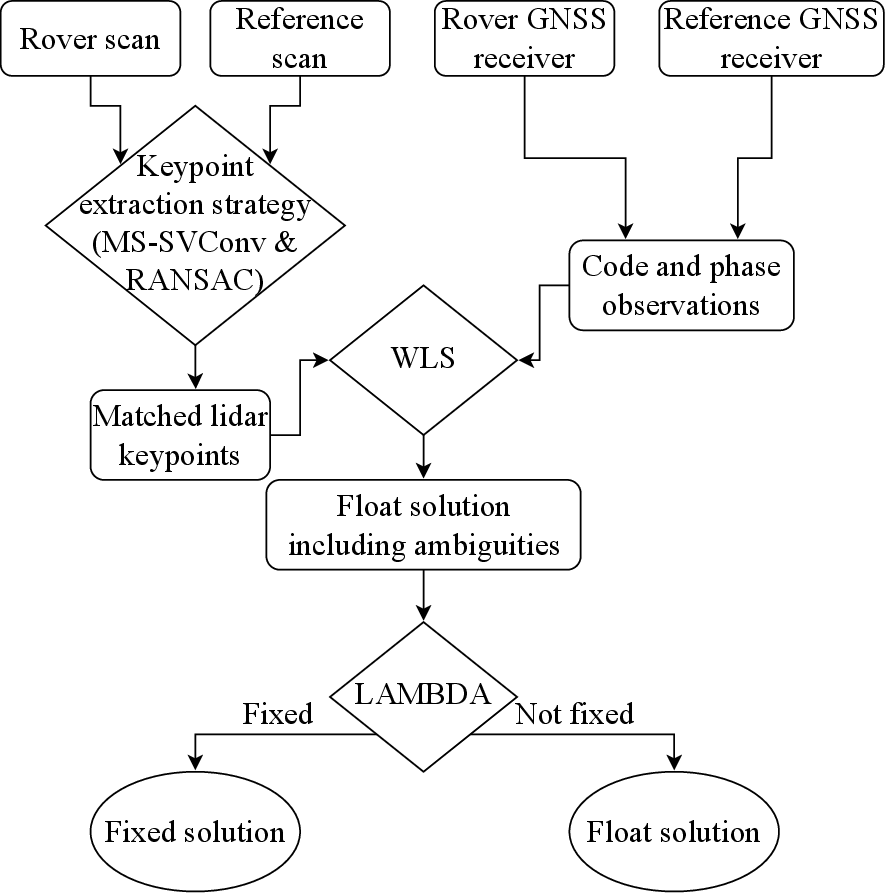}
    \caption{Overview of the proposed lidar-aided ambiguity resolution method.}
    \label{fig:workflow}
\end{figure}

The key to enabling GNSS carrier phase ambiguity resolution is to improve the precision of float solutions with redundant measurements. In this paper, lidar measurements are integrated with GNSS code observations by employing the Weighted Least Squares (WLS) method to obtain the float solutions. To achieve the `minimum-variance' float solutions, the weight matrix underlying WLS is taken as the inverse of the measurements' variance matrix~\cite{teun:adj}. The lidar measurements, in the form of 3D coordinates of matched keypoints, are produced by registering rover (online) scans collected by the laser scanner on-board a vehicle with reference (offline) scans obtained from a pre-built HD map. The corresponding keypoints are extracted to estimate a rigid transformation which provides the position of the origin of a rover scan, i.e., the vehicle, provided that the laser scanner is calibrated to align with the center of the vehicle. This lidar positioning method has been shown to be effective in terms of availability and accuracy of positioning in our previous work \cite{zhang2021seamless}. Fig.~\ref{fig:workflow} depicts the workflow of the proposed lidar-aided ambiguity resolution method, in which the lidar observation generation discussed in this section is reflected in the steps in the top left part.

Several coordinate systems are used throughout the following sections. We define \textit{e-frame} as the geocentric WGS84 frame, which is used for all GNSS measurements and the computations of the positioning solutions. The point clouds are originally collected in a Cartesian frame with the sensor at its center, namely the \textit{l-frame}. To integrate the generated lidar measurements with GNSS, reference scans in the HD map are transformed and aligned to a local frame \textit{c-frame} with an arbitrary origin first, then to the \textit{e-frame}.

\subsection{HD map definition}
An HD map includes information needed for vehicle positioning and can take various forms \cite{liu2020high}. In this research, we use an HD map that utilizes accurately georeferenced point clouds of the road environment collected from a previous time. Each point cloud is stored in its original \textit{l-frame} with the laser scanner as the origin, as well as a $4\times4$ georeferencing matrix to transform it to \textit{e-frame} so that the positioning solutions are in this coordinate system. Such point clouds that make up of the HD map are referred to as reference scans and must be available prior to the vehicle positioning tasks. A procedure for preparing an HD map using KITTI dataset \cite{Geiger2013IJRR} is provided in Section~\ref{sec:lidardata}.

\subsection{Learning-based keypoint extraction}
The registration between a pair of rover and reference scans requires correctly corresponded keypoints. For fast and accurate keypoint extraction, we use MS-SVConv, a multi-scale deep neural network that outputs feature vectors from point clouds while considering point density variation~\cite{horache2021mssvconv}. A model needs to be pre-trained on a large number of point clouds with ground truth alignments before it can be used to compute feature vectors.

Assuming that for one epoch a rover scan is collected by the lidar sensor and a nearby reference scan which shares an overlap is identified from the HD map using position estimates from less demanding techniques such as Standard Point Positioning (SPP), the pre-trained MS-SVConv model is applied to produce the feature vectors per point, based on which the keypoint matching is performed. Since keypoints extracted by MS-SVConv often contain incorrect correspondences, a subset of keypoints with their feature vectors are selected to register the two point clouds using Random Sample Consensus (RANSAC) \cite{fischler1981random}, which finds an outlier-free set of keypoint correspondences to estimate a transformation matrix that georeferences the rover scan from \textit{l-frame} to \textit{e-frame}. Positions of successfully matched keypoints are then used as lidar observations for vehicle positioning.

\subsection{Lidar observation equations}

The lidar observations are constructed by the estimation of the transformation from the rover scan to the reference scan, in which the translational parameters are identical to the vehicle position. The coordinates of the keypoints from the rover scan in \textit{l-frame} serve as the measurements, corresponding with the known coordinates of their matched counterparts from the reference scan in \textit{e-frame}. Thus, each keypoint provides 3 observations. For $n$ pairs of matched keypoints, the observation equation of lidar measurements is therefore as follows

\begin{equation}
    \label{eq:lidarobs}
    \bm{1}_{n\times1}\otimes\bm{b}+\bm{B}_L^{\top}[\bm{y}_L-\bm{e}_L]-\bm{c}=\bm{0}
\end{equation}

\noindent in which $\otimes$ denotes the Kronecker product \cite{henderson1983}, and $\bm{1}$ is a matrix of ones. $\bm{b}=[x,y,z]^{\top}$ is the $3\times1$ vector of the estimated vehicle position in \textit{e-frame} and $\bm{R}$ is a $3\times3$ matrix of the unknown rotational parameters. The Jacobian matrix is given by $\bm{B}_L^{\top}=\bm{I}_n\otimes\bm{R}$. By defining $\bm{y}_j=[x_{j,l},y_{j,l},z_{j,l}]^\top\,(j=1,\ldots,n)$ as the vector of the measured coordinates of one keypoint in \textit{l-frame} and concatenating all the keypoints, the $3n\times1$ vectors of measured keypoint coordinates, measurement residuals and known keypoint coordinates, namely $\bm{y}_L$, $\bm{e}_L$ and $\bm{c}$, are formed as

\begin{gather*}
    \bm{y}_L=[\bm{y}_1^\top,\ldots,\bm{y}_n^\top]^\top\\
    \bm{e}_L= [e_{x,1},e_{y,1},e_{z,1},\ldots,e_{x,n},e_{y,n},e_{z,n}]^\top\\
    \bm{c}=[x_{1,e},y_{1,e},z_{1,e},\ldots,x_{n,e},y_{n,e},z_{n,e}]^\top
\end{gather*}

For integration with GNSS observations in the WLS method which will be discussed in the next section, the uniform weight matrix of lidar measurements is defined by

\begin{equation}
    \bm{W}_L=\frac{1}{\sigma_L^2}\bm{I}_{3n}
\end{equation}

\noindent with the root mean squared residual distance \mbox{$\sigma_L=\sqrt{(\sum_{k=1}^{n}v_k^2)/n}$}, in which $v_k$ is the residual distance between a registered keypoint and its correspondence in the reference scan in the RANSAC estimation.

\section{GNSS-lidar integration and ambiguity resolution \label{sec:gnss}}

In this section, the model integrating lidar and double-differenced (DD) GNSS observations to deliver the float solutions including ambiguities and using them for IAR is discussed. It is also reflected in the top right and bottom parts of Fig.~\ref{fig:workflow}. We base our method on the assumptions that the collected lidar and GNSS measurements are time-synchronized and are both referenced from the center of the vehicle.

\subsection{Double-differenced GNSS observation equations}

Using $\Delta\bm{p},\,\Delta\bm{\phi}\in\mathbb{R}^{f(m-1)}$ to denote the vectors of DD observed-minus-computed code and carrier phase measurements for $m$ visible satellites and $f$ frequencies collected in one epoch, respectively, the two can be concatenated as $\bm{y}_G=[\Delta\bm{p}^\top,\Delta\bm{\phi}^\top]^\top$. Assuming that the baseline between the GNSS receiver installed on the vehicle and the reference station is short enough so that the DD atmospheric delays can be ignored, the linearized GNSS observation equation is given as \cite{khodabandeh2021study,khodabandeh2018impact}

\begin{equation}
    \label{eq:gnssobs}
    \bm{\bar{\Lambda}}\bm{a}+[\bm{1}_{2f\times1}\otimes\bm{I}_{m-1}]\bm{\rho}-[\bm{y}_G-\bm{e}_G]=\bm{0}
\end{equation}

\noindent where $\bm{\bar{\Lambda}}=[\bm{0}_{f(m-1)\times{f(m-1)}},\bm{\Lambda}\otimes\bm{I}_{m-1}]^\top$ with $f\times{f}$ diagonal matrix $\bm{\Lambda}=\mathrm{diag(}\lambda_1,\ldots,\lambda_f\mathrm{)}$ links the DD ambiguities $\bm{a}\in\mathbb{Z}^{f(m-1)}$ with the GNSS measurements by the wavelength $\lambda_t\,(t=1,\ldots,f)$ of each observed frequency. The DD satellite-to-receiver range vector $\bm{\rho}$ can be linearized as $\bm{\rho}-\bm{\rho}_o\approx \bm{G}\Delta\bm{b}$, with $\bm{\rho}_o$ being its approximate version. The coefficient matrix $\bm{G}=\bm{D}^{\top}\bm{\bar{G}}$ has the dimensions of $(m-1)\times{3}$, with the $m\times3$ matrix $\bm{\bar{G}}$ containing the satellite-to-receiver direction unit vectors, and the $m\times(m-1)$ matrix $\bm{D}$ forming the between-satellite differences. Therefore, $\Delta\bm{b}$ is a vector of the increments to the unknown vehicle position. Similar to the lidar observation equation (\ref{eq:lidarobs}), the vector $\bm{e}_G$ contains the residuals of the GNSS code and carrier phase measurements.

The GNSS measurements are weighted according to the satellite elevation angles $\theta_s\,(s=1,\ldots,m)$. The $m\times{m}$ dimensionless weight matrix for \textit{undifferenced} GNSS measurements can hence be constructed as

\begin{equation}
    \bm{W}_G=\mathrm{diag(}sin^2{\theta_1},\ldots,sin^2{\theta_m}\mathrm{)}
\end{equation}

\noindent Accordingly, the weight matrices of DD code and carrier phase measurements are given as $\bm{W}_p=\frac{1}{2\sigma^2_p}(\bm{D}^\top\bm{W}_G^{-1}\bm{D})^{-1}$ and 
$\bm{W}_\phi=\frac{1}{2\sigma^2_\phi}(\bm{D}^\top\bm{W}_G^{-1}\bm{D})^{-1}$, respectively, where $\sigma_p$ and $\sigma_{\phi}$ denote the zenith-referenced standard deviations of the undifferenced code and phase observations~\cite{langley2017introduction}. The phase observations are assumed to be 100 times more precise than their code counterparts in our implementation. Thus $\sigma_{\phi}^2 = \epsilon\, \sigma_{p}^2$, with $\epsilon=10^{-4}$ being the phase-to-code variance ratio. The factor $2$ in both expressions indicates that the variances of the measurements are doubled due to differencing.

\subsection{Float solution by GNSS-lidar integration}

With the observation equations (\ref{eq:lidarobs}) and (\ref{eq:gnssobs}) in place, which share the common unknown vehicle position $\bm{b}$, we can estimate the float solution using the WLS principle. Due to the intertwining of the unknown rotational parameters $\bm{R}$ and the measurements $\bm{y}_L$ in the lidar observation equations, the mixed measurement model is employed to perform the estimation~\cite{zhang2021seamless,teun:adj}. The mixed model combining measurements from the two sensors can be formed as follows

\begin{equation}
    \label{eq:mixedmodel}
    \begin{array}{c}
        \bm{f}(\bm{x},\bm{y}-\bm{e})=\bm{0}\,\stackrel{\mathrm{linearized}}{\Longrightarrow}\\
        \bm{w}-\bm{B}^{\top}\bm{e}+\bm{A}\Delta\bm{x}=\bm{0}
    \end{array}
\end{equation}

\noindent where the vector of unknown parameters $\bm{x}=[\bm{a}^\top,\bm{b}^\top,\bm{r}^\top]^\top$, with $\bm{r}$ being the vector form of a close approximation of $\bm{R}$, contains the DD ambiguities, the vehicle position and the rotational parameters of the rover scan. $\bm{y}=[\bm{y}_L^\top,\bm{y}_G^\top]^\top$ and $\bm{e}=[\bm{e}_L^\top,\bm{e}_G^\top]^\top$ are the vectors of measurements and residuals. The mixed model is linearized using the first-order Taylor expansion of $\bm{f}(\bm{x},\bm{y}-\bm{e})$ about the point $(\bm{x}_0,\bm{y})$ to give the expression on the right hand side of (\ref{eq:mixedmodel}), in which $\bm{x}_0$ is the approximated version of $\bm{x}$ in each iteration, giving the unknown increment vector $\Delta\bm{x}=\bm{x}-\bm{x}_0$. Hence, $\bm{w}=\bm{f}(\bm{x}_0,\bm{y})$. The Jocabian matrices $\bm{A}$ and $\bm{B}^\top$ for $\bm{x}$ and $\bm{y}$ are given as follows

\begin{equation}
    \label{eq:mixedABt}
        \bm{A}=
        \begin{bmatrix}
           \bm{0}  & \bm{A}_L\\
            \bm{\bar{\Lambda}} & \bm{A}_G
        \end{bmatrix},\quad
        \bm{B}^{\top}=\mathrm{blkdiag}(\bm{B}_L^{\top},-\bm{I}_{2f(m-1)})
\end{equation}

\noindent with $\mathrm{blkdiag(\cdot)}$ denoting a block diagonal matrix. The Jacobian sub-matrices are given by $\bm{A}_L=\left[\bm{1}_{n\times1}\otimes\bm{I}_3\quad\bm{L}\right]$, and $\bm{A}_G=\left[\bm{1}_{2f\times1}\otimes\bm{G} \quad\bm{0}\right]$, in which matrix  $\bm{L}=[\bm{y}_1\otimes\bm{I}_3,\ldots,\bm{y}_n\otimes\bm{I}_3]^\top$ is formed by the lidar measurements $\bm{y}_i$ ($i=1,\ldots,n$). Application of WLS to the mixed model (\ref{eq:mixedmodel}) gives the float increment solution $\Delta\bm{\hat{x}}$ and their variance matrix $\bm{Q}_{\bm{\hat{x}\hat{x}}}$ as

\begin{equation}
    \label{eq:LSsolution}
    \begin{aligned}
        & \bm{e}^\top\bm{W}\bm{e}\mapsto\mathrm{minimum}\quad\Longrightarrow\\
        & \left\{
            \begin{array}{l}
                \Delta\bm{\hat{x}}=-[\bm{A}^\top\bm{M}\bm{A}]^{-1}\bm{A}^\top\bm{M}\bm{w}\\
                \bm{Q}_{\bm{\hat{x}\hat{x}}}=[\bm{A}^\top\bm{M}\bm{A}]^{-1}
            \end{array}
        \right.
    \end{aligned}
\end{equation}

\noindent with $\bm{M}=[\bm{B}^\top\bm{W}^{-1}\bm{B}]^{-1}$. The weight matrix $\bm{W}$ is constructed as $\bm{W}=\mathrm{blkdiag}(\bm{W}_L,\bm{W}_p,\bm{W}_{\phi})$. The float solution is iteratively computed as $\bm{\hat{x}}=\bm{x}_0+\Delta\bm{\hat{x}}$, in which $\bm{x}_0$ is replaced by the solution from the previous iteration. The solution convergence is considered reached by a stopping criterion, for instance, when the magnitude of $\Delta\bm{\hat{x}}$ is smaller than a specified threshold.

\subsection{Integer ambiguity resolution}

The float solution obtained from (\ref{eq:LSsolution}) contains the estimated vehicle position ($\bm{\hat{b}}$), DD ambiguities which are real-valued ($\bm{\hat{a}}$) and the rotational parameters for the georeferencing of the rover scan ($\bm{\hat{r}}$). For simplicity, we use $\bm{g}=[\bm{b}^\top,\bm{r}^\top]^\top$ to denote the non-ambiguity unknown parameters here. In order to utilize the highly precise carrier phase measurements for positioning, the float ambiguities $\bm{\hat{a}}\in\mathbb{R}^{f(m-1)}$ need to be mapped to the correct integers $\bm{\check{a}}\in\mathbb{Z}^{f(m-1)}$ using an ambiguity resolution method to yield the fixed solution \cite{teunissen2017carrier}, provided that the estimated float parameters follow the normal distribution:

\begin{equation}
    \begin{bmatrix}
        \bm{\hat{a}}\\
        \bm{\hat{g}}
    \end{bmatrix}
    \sim
    \mathcal{N}
    \begin{pmatrix}
        \begin{bmatrix}
            \bm{a}\\
            \bm{g}
        \end{bmatrix}
        ,
        \begin{bmatrix}
            \bm{Q_{\hat{a}\hat{a}}} & \bm{Q_{\hat{a}\hat{g}}} \\
            \bm{Q_{\hat{g}\hat{a}}} & \bm{Q_{\hat{g}\hat{g}}}
        \end{bmatrix}
    \end{pmatrix}
\end{equation}

In this paper, we use the well known LAMBDA method for IAR, which employs the Integer Least-Squares (ILS) ambiguity estimator \cite{teunissen1995least}. By assessing the float $\bm{\hat{a}}$ and $\bm{Q_{\hat{a}\hat{a}}}$, LAMBDA outputs the fixed integer ambiguities $\bm{\check{a}}$, as well as the evaluated formal success rate \cite{verhagen2012lambda}. In order to determine whether the fixed ambiguities should be accepted, an acceptance test is needed. For example, the formal success rate can be required to be higher than a given threshold such as 99.9\%. Otherwise, the float solution is retained. Once $\bm{\check{a}}$ is accepted, the fixed solution of the remaining parameters, say $\bm{\check{g}}$ and its variance matrix $\bm{Q_{\check{g}\check{g}}}$, can be obtained by \cite{teunissen1995least} 

\begin{equation}
    \left\{
    \begin{array}{l}
        \bm{\check{g}}=\bm{\hat{g}}-\bm{Q_{\hat{g}\hat{a}}}\bm{Q_{\hat{a}\hat{a}}}^{-1}[\bm{\hat{a}}-\bm{\check{a}}]\\[6pt]
        \bm{Q_{\check{g}\check{g}}}=\bm{Q_{\hat{g}\hat{g}}}-\bm{Q_{\hat{g}\hat{a}}}\bm{Q_{\hat{a}\hat{a}}}^{-1}\bm{Q_{\hat{g}\hat{a}}}^{\top}
    \end{array}
    \right.
\end{equation}

\noindent where the first 3 elements of $\bm{\check{g}}$ are the fixed position $\bm{\check{b}}$.

\section{Theoretical performance assessment using ADOP\label{sec:analysis}}

So far we have discussed the model to integrate GNSS and lidar measurements for IAR. 
To provide a prediction of the method's IAR performance {\em before} taking any measurements, one needs the variance matrix of the float ambiguity solutions. Such variance matrix enables us to a priori evaluate the ambiguity success rate and quantify the IAR performance using its associated ADOP, which indicates the upper bound of the ambiguity success rate~\cite{verhagen2005reliability,teunissen1997canonical}. In this section, we therefore present closed-form expressions for both the ambiguity variance matrix and its ADOP. Given such closed-form analytical measures, we will then evaluate the formal ambiguity success rates and ADOP of the proposed lidar-aided model with the aid of Ps-LAMBDA software \cite{verhagen2013ps}.

\subsection{Variance matrix of the float solution}
Firstly, the expressions of the precision of float ambiguities and positioning solution are developed to support the evaluation. Applying the WLS principle yields the normal matrix of the estimated parameters for lidar as \cite{teun:adj,zhang2021seamless}

\begin{equation}
    \bm{N}_L=\bm{A}_{L}^{\top}\bm{M}_{L}\bm{A}_{L}=
    \begin{bmatrix}
        \bm{N}_{\bm{\hat{b}\hat{b}}} & \bm{N}_{\bm{\hat{b}\hat{r}}}\\[6pt]
        \bm{N}_{\bm{\hat{b}\hat{r}}}^{\top} & \bm{N}_{\bm{\hat{r}\hat{r}}}
    \end{bmatrix}
\end{equation}

\noindent with $\bm{M}_L=[\bm{B}_L^{\top}\bm{B}_L]^{-1}$. The reduced normal matrix of the estimated position is therefore $\bm{\bar{N}}_{\bm{\hat{b}\hat{b}}}=\bm{N}_{\bm{\hat{b}\hat{b}}}-\bm{N}_{\bm{\hat{b}\hat{r}}}\bm{N}_{\bm{\hat{r}\hat{r}}}^{-1}\bm{N}_{\bm{\hat{b}\hat{r}}}^{\top}$, which leads to the variance matrix of the lidar-only positioning solution as follows

\begin{equation}
    \label{eq:fQlidar}
    \bm{Q}_{\bm{\hat{b}\hat{b}},L}=\sigma_L^2\bm{\bar{N}}_{\bm{\hat{b}\hat{b}}}^{-1}=\sigma_L^2[\bm{N}_{\bm{\hat{b}\hat{b}}}-\bm{N}_{\bm{\hat{b}\hat{r}}}\bm{N}_{\bm{\hat{r}\hat{r}}}^{-1}\bm{N}_{\bm{\hat{b}\hat{r}}}^{\top}]^{-1}
\end{equation}

On the other hand, the variance matrix of the float solution obtained with GNSS code observations is given as

\begin{equation}
    \label{eq:fQgnss}
    \bm{Q}_{\bm{\hat{b}\hat{b}},G} = \frac{2\sigma_p^2}{f}\left[\bm{G}^{\top}\bm{W}_G\bm{PG}\right]^{-1}
\end{equation}

\noindent with the projection matrix $\bm{P}=\bm{W}_G^{-1}\bm{D}[\bm{D}^{\top}\bm{W}_G^{-1}\bm{D}]^{-1}\bm{D}^{\top}$.  Therefore, (\ref{eq:fQlidar}) and (\ref{eq:fQgnss}) can be combined to give the variance matrix of the integrated float solution:

\begin{equation}
    \label{eq:fQ}
    \bm{Q_{\hat{b}\hat{b}}} = \frac{2\sigma_p^2}{f}[\bm{G}^{\top}\bm{W}_{G}\bm{PG} +\frac{2\sigma_p^2}{f\sigma_L^2}\bm{\bar{N}
    }_{\bm{\hat{b}\hat{b}}}]^{-1}
\end{equation}

Likewise, the variance matrix of float ambiguities computed using both code and lidar measurements takes the following form
\begin{equation}
    \label{eq:Qa}
    \begin{array}{lcl}
        \bm{Q_{\hat{a}\hat{a}}} & = & [2\sigma_{\phi}^2\,\bm{\Lambda}^{-2}\otimes(\bm{D}^{\top}\bm{W}_G^{-1}\bm{D})]\\
        & & + [\bm{\Lambda}^{-1}\bm{1}_{f\times{f}}\bm{\Lambda}^{-1}\otimes \bm{D}^{\top}\bm{G}\,\bm{Q_{\hat{b}\hat{b}}}\bm{G}^{\top}\bm{D}]
    \end{array}
\end{equation}

\noindent in which one can replace $\bm{Q_{\hat{b}\hat{b}}}$ with (\ref{eq:fQgnss}) to obtain the variance matrix of the GNSS-only float ambiguities $\bm{Q}_{\bm{\hat{a}\hat{a}},G}$. The ambiguity variance matrix (\ref{eq:Qa}) can serve as input to evaluate formal ambiguity success rates.

\subsection{ADOP of lidar-aided ambiguity resolution}
Next to formal ambiguity success rates, one can also evaluate the ADOP of one's measurement model to assess the underlying IAR performance. ADOP is defined on the basis of the ambiguity variance matrix as follows \cite{teunissen1997canonical}

\begin{equation}
    \label{eq:adop}
    {\rm ADOP}=\sqrt{|\bm{Q_{\hat{a}\hat{a}}}|}^{\frac{1}{f(m-1)}}\quad [{\rm cycles}]
\end{equation}

\noindent where $|\cdot|$ denotes the determinant of a matrix.
The smaller the ADOP, the higher the ambiguity success rate becomes \cite{khodabandeh2021study,teunissen2000adop}. For a minimum success rate of 99.9\%, ADOP should be smaller than 0.12 cycles, whereas for 99\%, it should be smaller than 0.14 cycles \cite{odijk2008adop}. For the single-epoch GNSS-only model (\ref{eq:gnssobs}), ADOP can be expressed as \cite{odijk2008adop}:
\begin{equation}
    \label{eq:ADOPg}
    {\rm ADOP}^{\rm G} = \sqrt{2}\,w_o\, \dfrac{\sigma_\phi}{\bar{\lambda}} \left[1+\frac{1}{\epsilon}\right]^{\frac{3}{2f(m-1)}}
\end{equation}
\noindent with $\epsilon=(\sigma_\phi^2/\sigma_p^2)$, $\bar{\lambda}=\prod_{j=1}^f\lambda_j^{\frac{1}{f}}$ and $w_o = \left[\frac{\sum_{s=1}^m w_s}{\prod_{s=1}^m w_s}\right]^{\frac{1}{2(m-1)}}$, where $w_s$ is the diagonal entries of $\bm{W}_G$. In order to compare the ADOP of GNSS-only (\ref{eq:ADOPg}), denoted by ${\rm ADOP}^{\rm G}$, with that of the lidar-aided method, say ${\rm ADOP}^{\rm GL}$, one can express their ratio as follows (see Appendix)

\begin{figure}[!t]
    \centering
    \subfloat[]{\includegraphics[width=.9\linewidth]{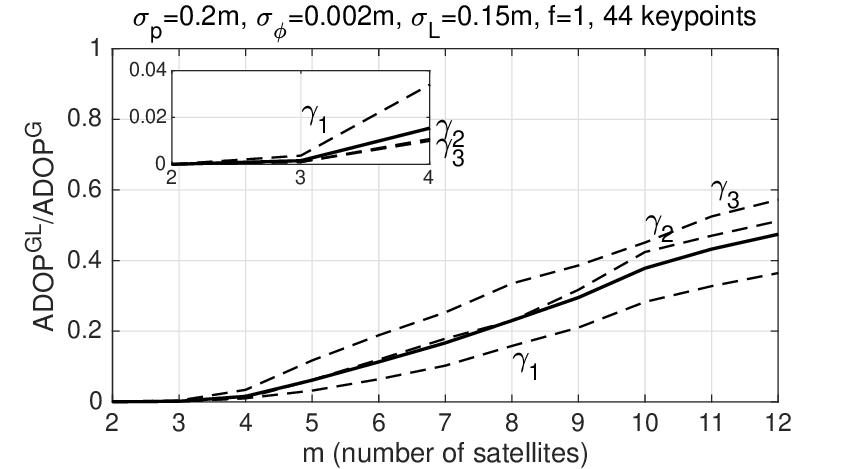}%
    \label{fig:ratio1}}
    \vfill
    \subfloat[]{\includegraphics[width=.9\linewidth]{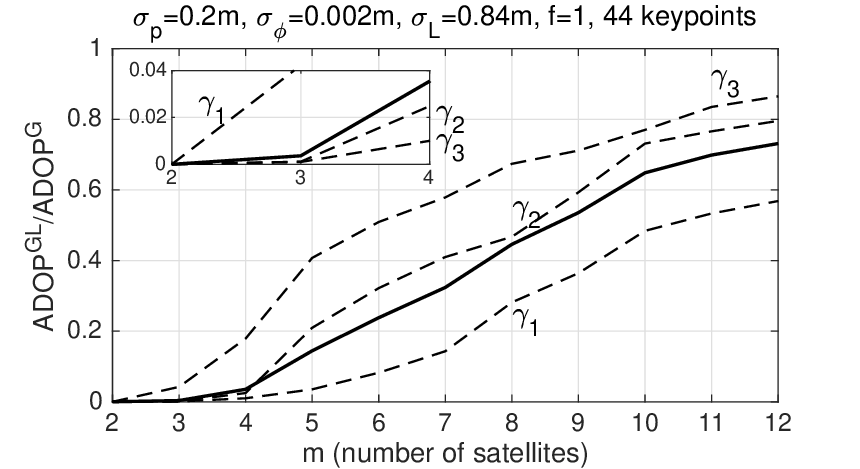}%
    \label{fig:ratio2}}
    \caption{ADOP-ratio (\ref{eq:ADOPratio}) (solid lines) and its three approximate versions (\ref{eq:ADOPratio2}) (dashed lines) as functions of the number of visible satellites $m$ for a single-frequency GNSS receiver ($f=1$), when 44 matched lidar keypoints are given. (a) ADOP-ratios with $\sigma_p=0.2$ m, $\sigma_{\phi}=0.002$ m, $\sigma_L=0.15$ m. (b) ADOP-ratios with $\sigma_p=0.2$ m, $\sigma_{\phi}=0.002$ m, $\sigma_L=0.84$ m.}
    \label{fig:ratio}
\end{figure}

\begin{equation}
    \label{eq:ADOPratio}
    \begin{array}{lcl}
    \dfrac{{\rm ADOP}^{\rm GL}}{{\rm ADOP}^{\rm G}} &=&|\bm{I}_3-\dfrac{1}{1+\epsilon}\bm{Q}_{\bm{\hat{b}\hat{b}},L}^{-1}\bm{Q_{\hat{b}\hat{b}}}|^{\frac{1}{2f(m-1)}}\vspace{1mm}\\
    &=&\prod\limits_{i=1}^3\left[1-\frac{1}{1+\epsilon}\frac{1}{\gamma_i}
    \right]^{\frac{1}{2f(m-1)}}
    \end{array}
\end{equation}
\noindent where the eigenvalues $\gamma_i$ ($i=1,2,3$) are the roots of the characteristic equations
\begin{equation}
    \label{eq:eigenvalues}
    |\bm{Q}_{\bm{\hat{b}\hat{b}},L} - \gamma_i\, \bm{Q_{\hat{b}\hat{b}}}| = 0,\quad i=1,2,3\vspace{2mm}\\
\end{equation}
The ADOP-ratio (\ref{eq:ADOPratio}) tells us the extent to which the ADOP of the lidar-aided model is smaller than that of the GNSS-only model given in (\ref{eq:ADOPg}). The first expression of (\ref{eq:ADOPratio}) indicates that the precision of the integrated solution  $\bm{Q}_{\bm{\hat{b}\hat{b}}}$ in (\ref{eq:fQ}) competes with that of the lidar-only solution $\bm{Q}_{\bm{\hat{b}\hat{b}},L}$ in (\ref{eq:fQlidar}) to reduce the ADOP-ratio. Consider the hypothetical scenario where the precision of the lidar measurements is extremely poor (i.e., $\bm{Q}_{\bm{\hat{b}\hat{b}},L}^{-1}\approx\bm{0}$), the ADOP-ratio reduces to 1, that is, ${\rm ADOP}^{\rm GL}\approx {\rm ADOP}^{\rm G}$. This would imply that the lidar measurements do not contribute to the integrated solutions. Now consider another extreme case where the lidar measurements are significantly more precise than the GNSS code measurements so that the precision of the integrated solutions becomes almost identical to that of the lidar-only solutions, i.e., $\bm{Q}_{\bm{\hat{b}\hat{b}}}\approx\bm{Q}_{\bm{\hat{b}\hat{b}},L}$ or $\bm{Q}_{\bm{\hat{b}\hat{b}},L}^{-1}\bm{Q}_{\bm{\hat{b}\hat{b}}}\approx\bm{I}_3$. Given the small phase-to-code variance ratio $\epsilon\approx 0$, such extreme case would therefore make the ADOP-ratio close to zero. This again makes sense, showing the more precise lidar measurements the smaller the ADOP-ratio (\ref{eq:ADOPratio}).

The second expression of (\ref{eq:ADOPratio}) reveals the link between the IAR performance of the GNSS-lidar integrated solution and the eigenvalues $1\leqslant \gamma_1\leqslant\gamma_2\leqslant\gamma_3$ defined in (\ref{eq:eigenvalues}). These eigenvalues are in fact the {\em stationary} values of the objective function $h(\bm{d}) =(\bm{d}^{\top}\bm{Q}_{\bm{\hat{b}\hat{b}},L}\bm{d})/(\bm{d}^{\top}\bm{Q_{\hat{b}\hat{b}}}\bm{d})$~\cite{teunissen1997canonical}, with $\bm{d}$ being an arbitrary unit direction vector. The smallest one, i.e., $\gamma_1$, indicates the minimum reduction in the variance of the lidar-only positioning solution when GNSS code data is integrated with the lidar data. Likewise, the largest eigenvalue $\gamma_3$ indicates the maximum reduction in the variance of the positioning solution. 

Each of these three eigenvalues can be served to `approximate' the ADOP-ratio (\ref{eq:ADOPratio}). For instance, by assuming that the eigenvalues are equal to one another, i.e. $\gamma_i=\gamma_k$ ($i\neq k$), three different approximate versions of (\ref{eq:ADOPratio}) are made as follows

\begin{equation}
    \label{eq:ADOPratio2}
    \dfrac{{\rm ADOP}^{\rm GL}}{{\rm ADOP}^{\rm G}} 
    \approx\left[1-\frac{1}{1+\epsilon}\frac{1}{\gamma_k}
    \right]^{\frac{3}{2f(m-1)}},\quad k=1,2,3
\end{equation} 

Fig.~\ref{fig:ratio} presents the ADOP-ratio (\ref{eq:ADOPratio}) (solid lines) and its three approximate versions (\ref{eq:ADOPratio2}) (dashed lines) as functions of the number of visible satellites $m$ when a {\em single-frequency} GNSS receiver is aided with 44 correctly matched lidar keypoints, which is the empirical minimum number of keypoints per epoch found in our experiment (cf. Section~\ref{sec:results}). The phase-to-code variance ratio $\epsilon$ is assumed to be $10^{-4}$. In Fig.~\ref{fig:ratio1}, precise lidar measurements are considered with $\sigma_L=0.15$ m, which is the mean value obtained from our experiment, whereas Fig.~\ref{fig:ratio2} shows the results for less precise lidar measurements with $\sigma_L=0.84$ m, which is approximately the maximum point spacing in point clouds used in the experiment collected by a Velodyne HDL-64E scanner~\cite{velodyne64}. In either case, it is observed that the ADOP-ratio (\ref{eq:ADOPratio}) gets closer to 1 the higher the number of satellites, meaning that the lidar-integration cannot contribute to the GNSS-only IAR performance by much when the receiver tracks a rather large number of satellites (e.g., $m\geqslant12$). Interestingly however, the close-to-zero ADOP-ratios for $m\leqslant5$ show that the lidar-integration can be indeed instrumental when not too many satellites are tracked. This is often the case in urban canyons where GNSS receivers frequently lose the tracking of GNSS signals. For the cases where standalone GNSS-only IAR is not possible, i.e. when $m<4$,  the integrated solution becomes almost identical to that of the lidar-only solutions, i.e., $\bm{Q}_{\bm{\hat{b}\hat{b}}}\approx\bm{Q}_{\bm{\hat{b}\hat{b}},L}$. That is why the ADOP-ratio is shown to be almost zero in the figure. In the following, this notion will be made more precise by comparing the underlying ADOPs and ambiguity success rates of the GNSS-only and lidar-aided models under various scenarios.

\subsection{Ambiguity resolution performance compared}

\begin{figure}[!t]
    \centering
    \subfloat[]{\includegraphics[width=.9\linewidth]{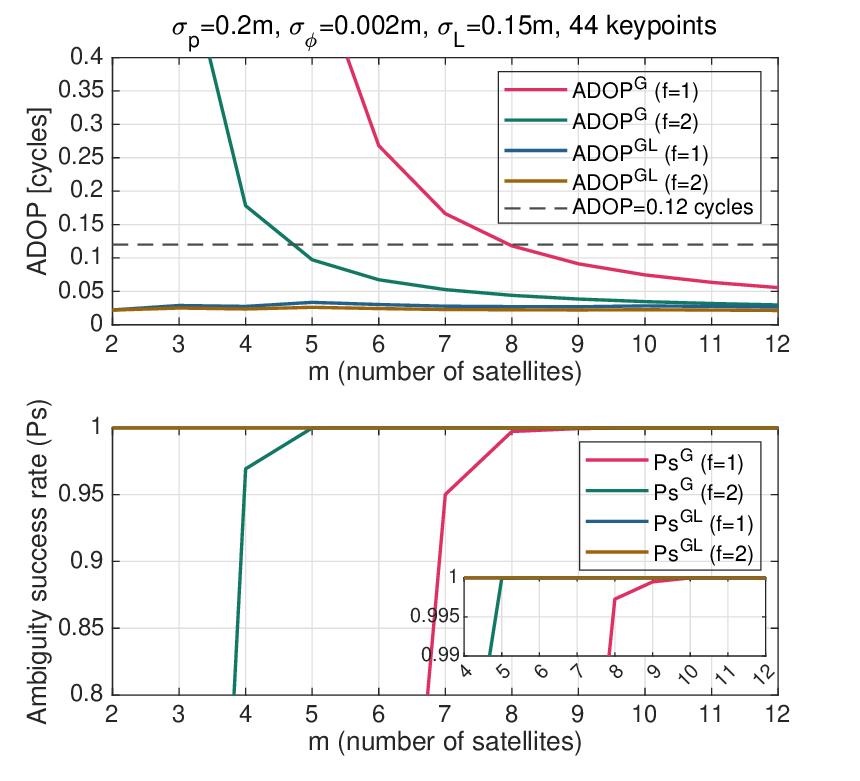}%
    \label{fig:ADOP_Ps_m_44}}
    \vfill
    \subfloat[]{\includegraphics[width=.9\linewidth]{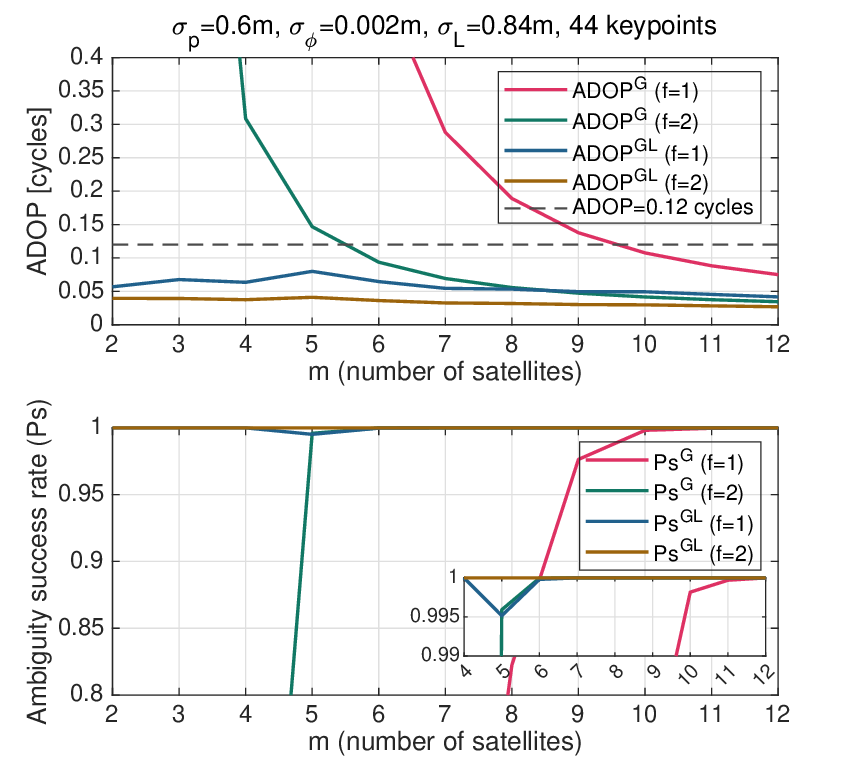}%
    \label{fig:ADOP_Ps_m_44p}}
    \caption{ADOP and ambiguity success rates of GNSS-only and lidar-aided models against different numbers of satellites. (a) ADOP and ambiguity success rates with $\sigma_p=0.2$ m, $\sigma_{\phi}=0.002$ m, $\sigma_L=0.15$ m. (b) ADOP and ambiguity success rates with $\sigma_p=0.6$ m, $\sigma_{\phi}=0.002$ m, $\sigma_L=0.84$ m.}
    \label{fig:ADOP_Ps}
\end{figure}

We now analyze the IAR performance using (\ref{eq:ADOPg}) and (\ref{eq:ADOPratio}), as well as the ambiguity success rates under various configurations for both the GNSS-only and lidar-aided models to study how many satellites they require for successful IAR. We first present and discuss the ambiguity resolution performance of the GNSS-only model in a generalized scenario without considering satellite elevation angles by simplifying $w_s$ to equal weights, i.e., $w_0=m^{\frac{1}{2(m-1)}}$. The precision of undifferenced code and phase data, namely $\sigma_p$ and $\sigma_\phi$, are assumed to be 0.2 m and 0.002 m, respectively, for high-grade receivers. Hence, the precision of phase observations is about 1\% of their wavelengths ($\sigma_\phi/\bar{\lambda}\approx10^{-2}$). Fig.~\ref{fig:ADOP_Ps} depicts the ADOP and ambiguity success rates (evaluated by Ps-LAMBDA, denoted by $Ps$) computed in different setups with respect to the number of tracked satellites. It is shown in Fig.~\ref{fig:ADOP_Ps_m_44} that with dual-frequency GNSS-only data ($f=2$), $\mathrm{ADOP^G}$ and the corresponding success rate can reach 0.097 cycles and 99.9\% with only 5 satellites, meaning that successful ambiguity resolution with single-system, dual-frequency observations is indeed possible {\em if} the corresponding GNSS code data are not imprecise (e.g., $\sigma_p \leqslant 0.6$ m). In this case, there is no need for the inclusion of lidar data. In comparison, when only single-frequency data from 5 satellites is available ($f=1$) for the GNSS-only model, $\mathrm{ADOP^G}$ and the success rate are evaluated as about 0.547 cycles and 11.2\%, respectively, which are insufficient for IAR. In order to constrain $\mathrm{ADOP^G}$ to 0.12 cycles, $m$ has to be increased to 8, while Ps-LAMBDA only reports a success rate higher than 99.9\% when $m=9$.

Note that the use of equal weights in this approximation ignores the elevation-dependent effects on the GNSS signals, an assumption which cannot always hold in practice. This highlights that the ambiguity resolution performance can be expected to be even poorer in practice. Moreover, low-grade GNSS receivers can have much less precise code observations. For example, assuming that code data is severely affected by the low quality of the receiver and/or antenna by setting $\sigma_p=0.6$ m and the same $\sigma_\phi$ as above, $\mathrm{ADOP^G}$ using single-frequency data becomes 1.247 cycles for 5 satellites and at least 10 satellites are needed for $\mathrm{ADOP^G}\leqslant0.12$ cycles, as illustrated in Fig.~\ref{fig:ADOP_Ps_m_44p}. To ensure an evaluated success rate of 99.9\%, at least 11 satellites are required. Therefore, it is not feasible to pursue instantaneous IAR using the GNSS-only model with single-frequency data, unless a large number of satellites from multiple systems can be tracked.

We now examine the lidar measurements for the float solutions to show their impact on ambiguity resolution. Similar to the optimistic configuration above, with $\sigma_p=0.2$ m and $\sigma_{\phi}=0.002$ m, each lidar observation is assumed to be slightly more precise than a code observation with $\sigma_L=0.15$ m. Using the same number of keypoints as that in Fig.~\ref{fig:ratio}, and $\gamma_i$ ($i=1,2,3$) obtained using (\ref{eq:eigenvalues}), $\mathrm{ADOP^{GL}}$ and $Ps$ with respect to the number of satellites are also presented in Fig.~\ref{fig:ADOP_Ps_m_44}. It is indicated that the integrated data consistently achieves similar or even better ambiguity resolution performance than dual-frequency GNSS-only data, with the two quantities always being around 0.02 cycles and 100\%. This advantage is more evident when there are only a few tracked satellites, as lidar becomes the main contributor to the float solutions. Notably, in the case of tracking only 2 or 3 satellites, the GNSS-only model fails to resolve integer ambiguities due to the lack of measurements, whereas the lidar-aided model can still enable IAR since the lidar measurements devote to the estimation of the three positional unknown parameters. Furthermore, as lidar provides a large number of measurements, the difference between the ambiguity resolution performances of using single-frequency or dual-frequency GNSS data becomes negligible.

On the other hand, lidar measurements can suffer from observational noise, environmental changes, dynamic objects, etc., leading to a lower precision. In a pessimistic scenario with $\sigma_L=0.84$ m, as well as $\sigma_p=0.6$ m for imprecise code measurements, $\mathrm{ADOP^G}$, $\mathrm{ADOP^{GL}}$ and their corresponding $Ps$ are shown in Fig.~\ref{fig:ADOP_Ps_m_44p}. Notably, due to the lower precision of code observations, the IAR performance of GNSS-only model decreases, requiring more satellites to be tracked for an ambiguity success rate of 99.9\%. In contrast, the lidar-aided counterpart is similar to previous results, with $\mathrm{ADOP^{GL}}$ and $Ps$ consistently being around 0.05 cycles and 99.9\%, suggesting that successful ambiguity resolution can still be achieved with any number of satellites (from 2) and/or frequencies under such pessimistic assumptions.

\subsection{ADOP evaluation of lidar-aided ambiguity resolution\label{sec:ADOPeval}}

\begin{figure}[!t]
    \centering
    \includegraphics[width=.7\linewidth]{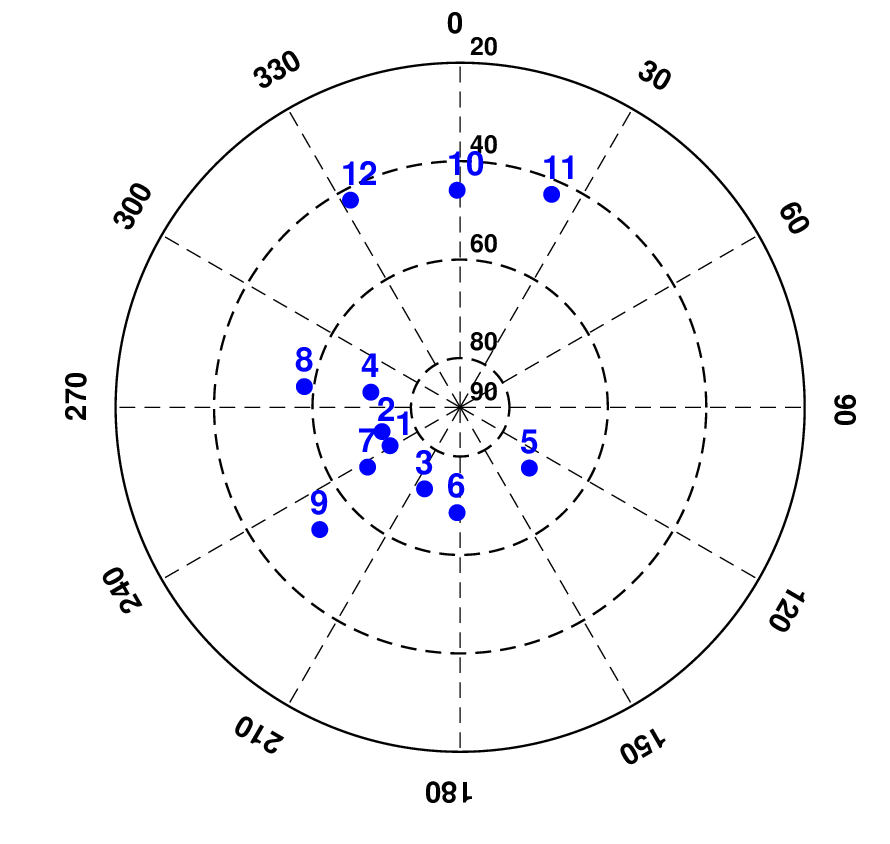}%
    \label{fig:skyplothigh}
    \caption{Skyplot of satellites in analyzed elevation-weighted scenario, satellite elevation angles are above 40{\degree}.}
    \label{fig:skyplot}
\end{figure}

So far we have learned that with the tracking of a sufficient number of GNSS satellites, IAR can be realized without lidar contribution, and the same holds true when there are limited GNSS observations but a number of lidar keypoints. To study the requirement of lidar precision for successful IAR, which is the other factor affecting the performance of the lidar-aided model, let us now consider the situation in which the number of corresponding keypoints is either 4 or 44, which are the theoretical minimum number to make (\ref{eq:lidarobs}) solvable and the empirical minimum used in previous comparison, respectively. $\sigma_p$ and $\sigma_\phi$ are maintained as $0.2$ m and $0.002$ m in this analysis. In contrast to the previous investigation, this analysis is established in an elevation-weighted scenario using GNSS satellites with high elevation angles to study the ambiguity resolution performance for vehicle positioning in urban canyons, where satellites with low elevation angles can be blocked by buildings. The satellite skyplot in Fig.~\ref{fig:skyplot} indicates the geometry of the satellites used in this analysis with respect to the receiver, where the satellite elevation angles are above 40{\degree}. These satellites are included by descending order of their elevation angles for the evaluation. Since the IAR performance can be sensitive to the geometric distributions of the keypoints when their number is low, each ADOP value is computed as the average of 100 trials with keypoints randomly selected around the receiver.

\begin{figure*}[!t]
    \centering
    \subfloat[]{\includegraphics[width=.33\linewidth]{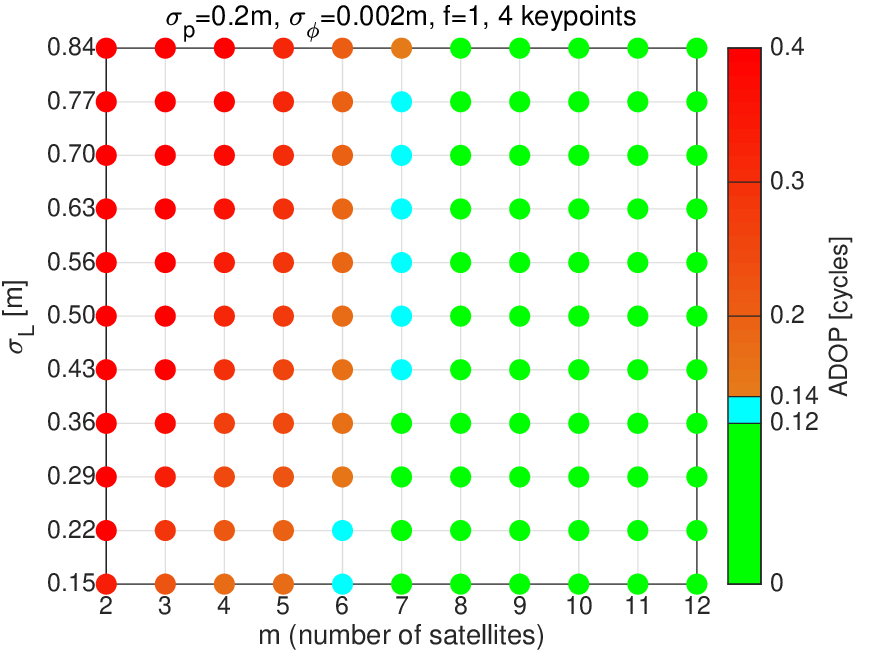}%
    \label{fig:ADOP_eval_high_f1_n4}}
    \subfloat[]{\includegraphics[width=.33\linewidth]{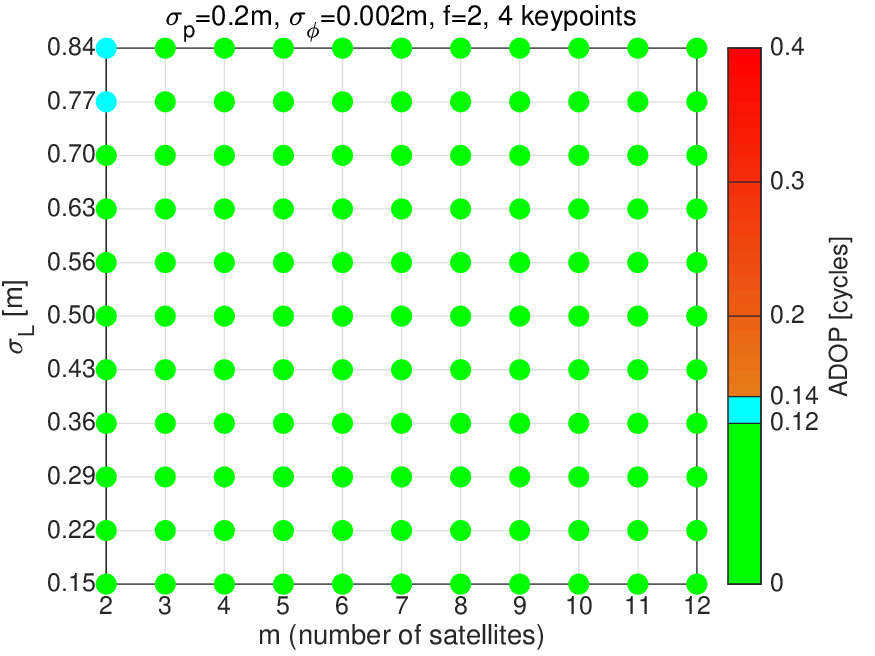}%
    \label{fig:ADOP_eval_high_f2_n4}}
    \subfloat[]{\includegraphics[width=.33\linewidth]{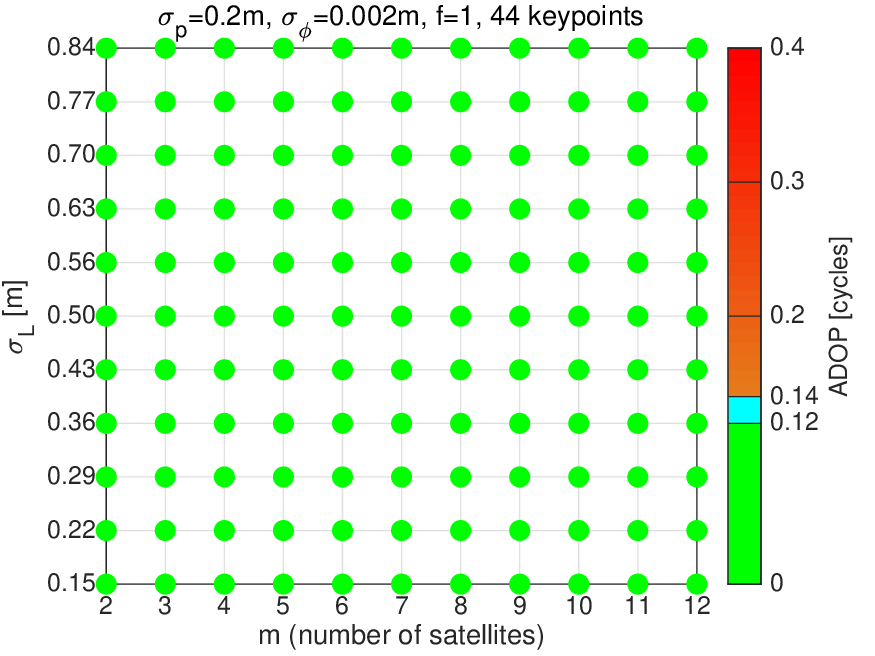}%
    \label{fig:ADOP_eval_high_f1_n44}}
    \caption{ADOP evaluation results for for different combinations of the number of satellites and lidar precision. (a) ADOP evaluation results with $f=1$ and 4 correctly matched keypoints; (b) ADOP evaluation results with $f=2$ and 4 correctly matched keypoints; (c) ADOP evaluation results with $f=1$ and 44 correctly matched keypoints.}
    \label{fig:ADOP_eval_high}
\end{figure*}

Fig.~\ref{fig:ADOP_eval_high} presents the $\mathrm{ADOP^{GL}}$ values for numerous combinations of the numbers of satellites ($m$) and lidar precision ($\sigma_L$) for the lidar-aided instantaneous ambiguity resolution method. For comparison, to obtain $\mathrm{ADOP^G}$ values below 0.12 cycles computed using (\ref{eq:ADOPg}), the GNSS-only model requires at least 9 high-elevation satellites for single-frequency GNSS data ($f=1$), which can be difficult to access in densely built-up areas for vehicle positioning, while 5 high-elevation satellites with dual-frequency GNSS data ($f=2$) are needed to produce the similar $\mathrm{ADOP^G}$. With the lidar-aided model, on the other hand, by using single-frequency GNSS observations and the theoretical minimum of 4 keypoints, the number of high-elevation satellites needed for the $\mathrm{ADOP^{GL}}$ value of 0.12 cycles reduces to 7, provided that $\sigma_L\leqslant0.36$ m, as shown in Fig.~\ref{fig:ADOP_eval_high_f1_n4}. Similarly, the lidar precision requirement can be relaxed to 0.77 m for $\mathrm{ADOP^{GL}}$ $\leqslant$ 0.14 cycles or the success rate upper bound of 99\%. It is therefore evident that 4 matched keypoints only can already reduce the required satellites by 2 for successful single-frequency IAR. In addition, Fig.~\ref{fig:ADOP_eval_high_f2_n4} shows that by introducing dual-frequency GNSS observations, $\mathrm{ADOP^{GL}}$ is always below 0.12 cycles for 3 and more satellites, whereas for 2 satellites, $\mathrm{ADOP^{GL}}$ would increase above 0.12 cycles but is still below 0.14 cycles for poor lidar precision ($\sigma_L\geqslant0.77$ m). Nonetheless, it is observed in Fig.~\ref{fig:ADOP_eval_high_f1_n44} that once the number of keypoints increases to the empirical minimum of 44, $\mathrm{ADOP^{GL}}$ can be kept below 0.12 cycles for any number of satellites from 2 even with the lowest lidar precision $\sigma_L=0.84$ m, which corresponds with the results in Fig.~\ref{fig:ADOP_Ps_m_44p}. It should be remarked that using the proposed keypoint extraction strategy, the quantity of corresponded keypoints can be expected to be at least equal to the empirical minimum upon successful registration. Conceivably, lidar-aided IAR is even more accessible by using dual-frequency GNSS data with this configuration, hence the results demonstrating the $\mathrm{ADOP^{GL}}$ values for dual-frequency GNSS observations and 44 correctly matched keypoints are omitted. In summary, the theoretical ADOP analysis has established that even with the theoretical minimum number of keypoints (i.e., 4), successful IAR is possible for any number of satellites using the lidar-aided method when dual-frequency GNSS data is available.
More importantly, when a reasonable amount of lidar keypoints are present (e.g., 44), the proposed method can substantially improve the ambiguity resolution performance to the extent that even single-frequency IAR is made feasible for only a few high-elevation satellites without requiring highly precise lidar measurements. This is a great advantage for vehicle positioning since the GNSS-only model can fail IAR in urban canyons due to restricted satellite visibility.

\section{Experimental setup \label{sec:setup}}

The proposed instantaneous lidar-aided ambiguity resolution method is evaluated in an experiment simulated using GNSS and lidar measurements from two real datasets. In this section, the data collection and pre-processing details are presented.
To verify the performance predictions made earlier in Section~\ref{sec:analysis},
we collected GNSS data from a 30-minute session of observations on a stationary point in a controlled environment, while the lidar data was obtained from the KITTI dataset \cite{Geiger2013IJRR} and simulated around the same point to build the HD map.

\subsection{GNSS data collection \label{sec:gnssdata}}
\begin{figure}[!t]
    \centering
    \includegraphics[width=.7\linewidth]{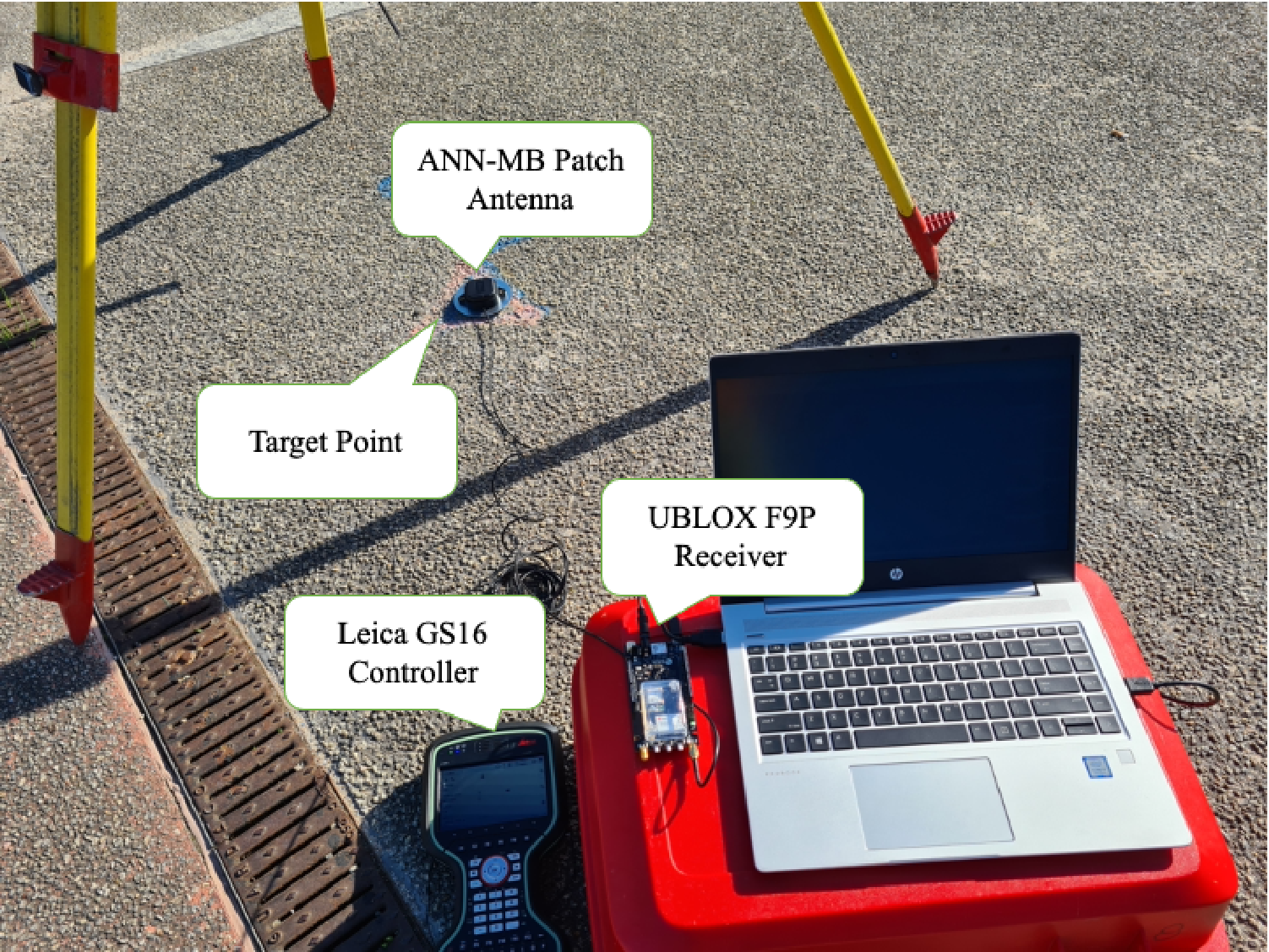}
    \caption{Equipment used for GNSS data collection, including \mbox{ANN--MB} patch antenna, u-blox F9P receiver and Leica GS16.}
    \label{fig:receiver}
\end{figure}

The GNSS raw observations used in the experiment were collected using a low-cost u-blox F9P dual-frequency receiver with an ANN--MB patch antenna between 1:56:29 AM and 2:26:28 AM, GPST (GPS time) on 29 June 2021, at the sampling rate of 1 Hz. The antenna was placed on a fixed point in an open-sky environment in Melbourne, Australia. A survey-grade GNSS receiver Leica GS16 was also used to measure the same point simultaneously to provide the ground truth coordinates as reference. From now on, this point will be referred to as Target. The equipment configuration is shown in Fig. \ref{fig:receiver}.

\begin{figure}[!t]
    \centering
    \includegraphics[width=.7\linewidth]{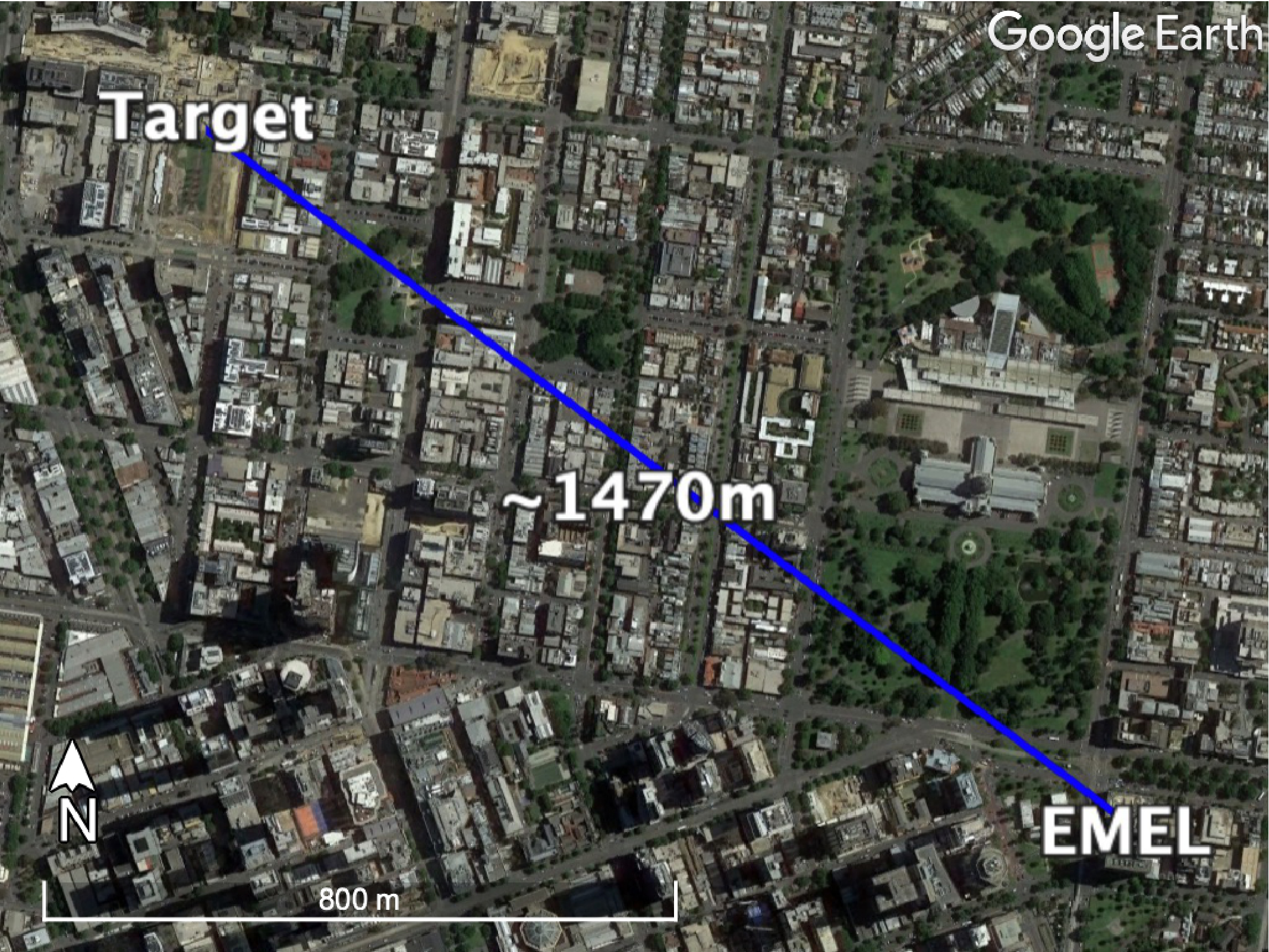}
    \caption{Target--EMEL baseline in Melbourne, Australia. Left: Target point; middle: baseline configuration; right: EMEL CORS, basemap from Google, 2018.}
    \label{fig:baseline}
\end{figure}

For data obtained from both receivers, differential code and phase observations are derived using a nearby Continuously Operating Reference Station (CORS), namely EMEL, which is equipped with a Trimble NETR9 receiver. Therefore, a short baseline \mbox{Target--EMEL} illustrated in Fig. \ref{fig:baseline} is formed with the approximate distance of 1470 m. By defining \textit{empirical success rate} as the proportion of positioning epochs with correctly fixed integer ambiguities, we take the DD ambiguities computed using GPS+QZSS (L1/L2) solutions obtained with u-blox F9P and EMEL as the benchmark ambiguities to evaluate the IAR capability of the proposed method, as the formal ambiguity success rates are assessed higher than 99.9\% for all epochs using this configuration.

\subsection{KITTI lidar data pre-processing \label{sec:lidardata}}

In Section \ref{sec:lidar} we presented the process of producing lidar measurements by learning-based point cloud registration. Here we provide the pre-processing of lidar data obtained from the KITTI dataset for simulating the pre-built and georeferenced HD map in \textit{e-frame}, so that the generated lidar measurements can be integrated with their GNSS counterparts. A total of 3600 point clouds collected in Karlsruhe, Germany with Velodyne HDL-64E are split into equal numbers of rover and reference scans. In other words, we assume that the rover scans are collected by the lidar sensor on-board of the rover vehicle, whereas the reference scans are used for the HD map. Due to the lack of SPP positions in the dataset, the rover-reference scan matches are pre-assigned with a 3 s time interval to ensure that each pair of rover and reference scans contains a reasonable amount of overlap to enable registration.


The point clouds are originally in \textit{l-frame}, they need to be transformed to \textit{c-frame}, which is equivalent to the \textit{l-frame} of the first scan in the sequence, then to \textit{e-frame} to produce lidar measurements that can be combined with their GNSS counterparts. For the $i\textrm{th}$ ($i=1,2,\ldots$) point cloud in the sequence, the $4\times4$ transformation $\bm{T}_{i,l}^{c}$ which transforms it to \textit{c-frame} is given in the ground truth poses of KITTI. Hence, for each pair of rover and reference scans we have transformations $\bm{T}_{rov,l}^{c}$ and $\bm{T}_{ref,l}^{c}$, and the transformation to align the two can be obtained as

\begin{equation}
    \bm{T}_{ref,l}^{rov,l}={\bm{T}_{rov,l}^{c}}^{-1}\cdot\bm{T}_{ref,l}^{c}
\end{equation}

Note that this transformation is not used for positioning, but rather simulating the reference scans at locations near Target so that the estimated vehicle positions should coincide with Target. A transformation matrix $\bm{T}_{rov,l}^e$ that georeferences the rover scan is defined to align it with the ground truth coordinates of Target measured by the survey-grade receiver. The reference scan is therefore georeferenced in \textit{e-frame} using: 



\begin{equation}
    \label{eq:georef}
    \bm{T}_{ref,l}^e=\bm{T}_{rov,l}^{e}\cdot\bm{T}_{ref,l}^{rov,l}
\end{equation}

\noindent which is applied to the matched keypoints found in the reference scans upon successful registration so that the computed lidar measurements are in \textit{e-frame}. In the next sections, the positioning performance will be evaluated in the local East-North-Up (ENU) directions. Due to this simulation of lidar data, the vehicle's body frame will align with the local ENU frame (i.e., along-the-road direction corresponds with Easting while across-the-road direction corresponds with Northing).
\section{Results \label{sec:results}}

In this section, we present the experimental results in terms of positioning accuracy and IAR performance of the proposed lidar-aided ambiguity resolution method. Using the GNSS and lidar data prepared in Section \ref{sec:setup}, the proposed method is tested by resolving the position of Target for 1800 epochs using \mbox{GPS (L1)+lidar} data to simulate the limited satellite visibility in urban environments. In addition, other positioning approaches using GPS (L1), GPS (L1/L2), GPS+QZSS (L1/L2) and Lidar-only measurements are used for comparison. In terms of the precision of GNSS code and carrier phase measurements, $\sigma_p$ and $\sigma_\phi$ are chosen as 0.2 m and 0.002 m, respectively, as the GNSS data was collected in a controlled environment despite a low-cost receiver was used. On the other hand, $\sigma_L$ for the weighting of lidar measurements is computed as approximately 0.15 m on average, therefore $\sigma_p>\sigma_L\gg\sigma_\phi$.

\subsection{Keypoint matching accuracy}

\begin{figure}[!t]
    \centering
    \includegraphics[width=.8\linewidth]{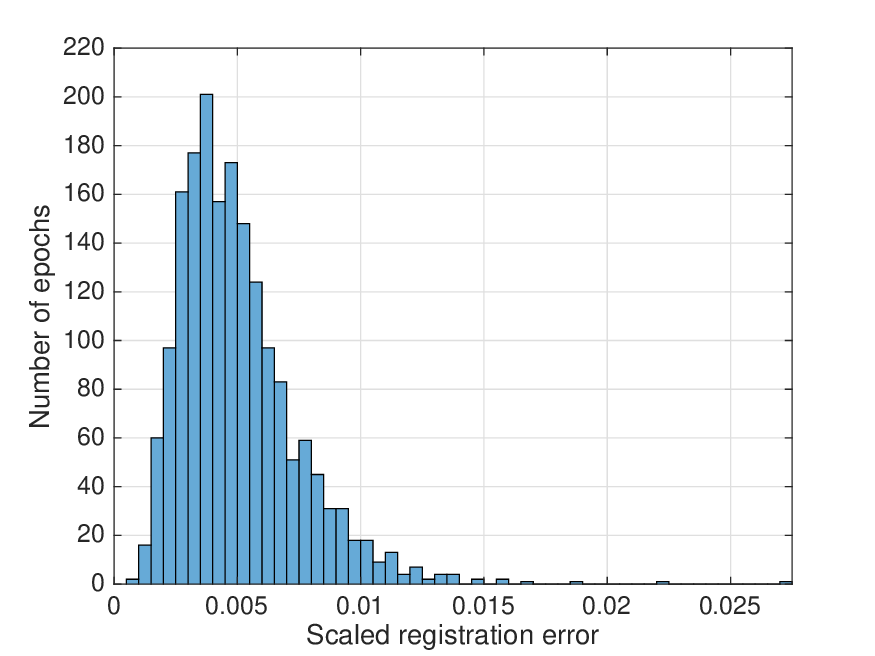}
    \caption{Distribution of SRE values of the tested scans registered using MS-SVConv.}
    \label{fig:sre}
\end{figure}

Keypoint matching to produce lidar measurements is the first step of the proposed positioning method. Feature vectors are computed for all points in the scans using MS-SVConv and 3000 pairs of them per epoch are randomly selected to estimate the transformation to align the rover scan to the reference scan using RANSAC \cite{fischler1981random}. Instead of training and testing the deep learning model on the same dataset, we use the model pre-trained with ETH dataset\cite{pomerleau2012challenging} and perform inference on the KITTI lidar data to test the transferability of MS-SVConv. In order to numerically evaluate the accuracy of the matched keypoints, we use SRE (scaled registration error)\cite{fontana2021benchmark}, a scalar measure of registration accuracy which considers both rotational and translational errors. For registered point cloud $\bm{P}$ and its ground truth counterpart $\bm{U}$ with $q$ corresponded points, namely $\bm{p}_k\in\bm{P}$ and $\bm{u}_k\in\bm{U}$, given that $\bm{\bar{p}}$ is the geometric centroid of $\bm{P}$, the SRE is computed with

\begin{equation}
    \label{eq:sre}
    SRE(\bm{P},\bm{U})=\frac{1}{q}\sum_{k=1}^{q}\frac{\Vert\bm{p}_k-\bm{u}_k\Vert}{\Vert\bm{p}_k-\bm{\bar{p}}\Vert}
\end{equation}

\noindent where $\Vert\cdot\Vert$ denotes the Euclidean norm. Since the registration error of each point is scaled by its distance to the centroid, one can infer that SRE reflects the phenomenon that rotational errors have a larger impact on points further from the laser scanner. As a result, registration is successfully conducted for all of the 1800 pairs of rover and reference scans, averaging 134 keypoints per epoch, with the minimum number of 44. Fig.~\ref{fig:sre} shows the distribution of the SRE values, in which 1044 epochs have SRE smaller than 0.005, while the maximum is 0.027. In comparison, Fontana \textit{et al.} \cite{fontana2021benchmark} obtained a mean SRE of 0.408 using the Iterative Closest Point (ICP) algorithm on multiple public datasets. Therefore, all of the registrations are considered accurate, and MS-SVConv exhibits a good transferability on point clouds from different environments.

\subsection{Positioning accuracy}

To demonstrate the accuracy of the tested positioning approaches, we compute the RMSE (root mean squared error) of the solutions with respect to the ground truth. Lidar-only positioning solutions are produced using the generated lidar measurements since the estimated translational parameters are equivalent to the sought unknown positions. For GNSS-involved methods, we use an acceptance test that integer ambiguities are fixed when the formal ambiguity success rate is evaluated as equal to or above 99.9\%, otherwise the float solutions are retained.

\begin{table}[!t]
    \caption{Horizontal, Vertical and 3D RMSE of Positioning Solutions from All Tested Methods (ambiguities are fixed when $Ps\geqslant99.9\%$). \label{tab:posres}}
    \begin{center}
        \begin{tabular}{cccc}
            \hline
            Positioning & Horizontal & Vertical & 3D \\
            Method & RMSE [m] & RMSE [m] & RMSE [m] \\
            \hline
            GPS (L1) & 0.648 & 1.190 & 1.355 \\
            \hline
            GPS (L1/L2) & 0.052 & 0.065 & 0.083 \\
            \hline
            GPS+QZSS (L1/L2) & 0.009 & 0.018 & 0.020 \\
            \hline
            Lidar-only & 0.034 & 0.017 & 0.038 \\
            \hline
            GPS (L1)+lidar & 0.008 & 0.014 & 0.016  \\
            \hline
        \end{tabular}
    \end{center}
\end{table}

\begin{figure*}[!t]
    \centering
    \subfloat[]{\includegraphics[width=.33\linewidth]{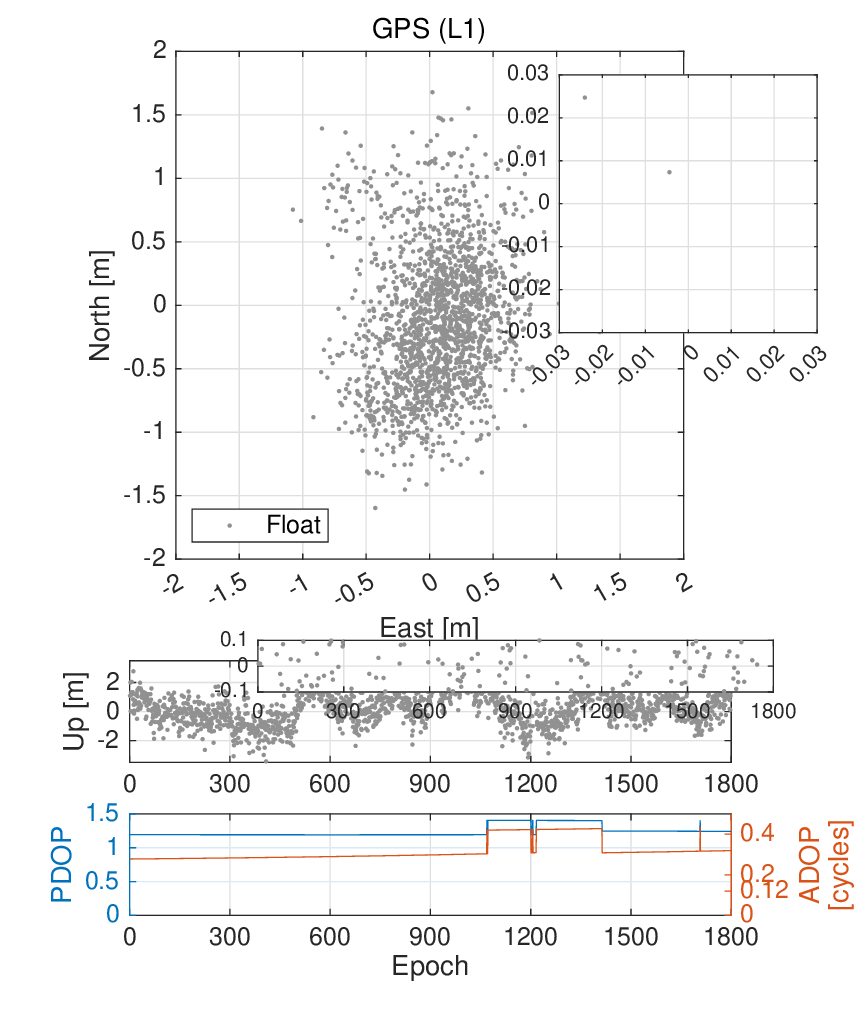}%
    \label{fig:res_gps}}
    \subfloat[]{\includegraphics[width=.33\linewidth]{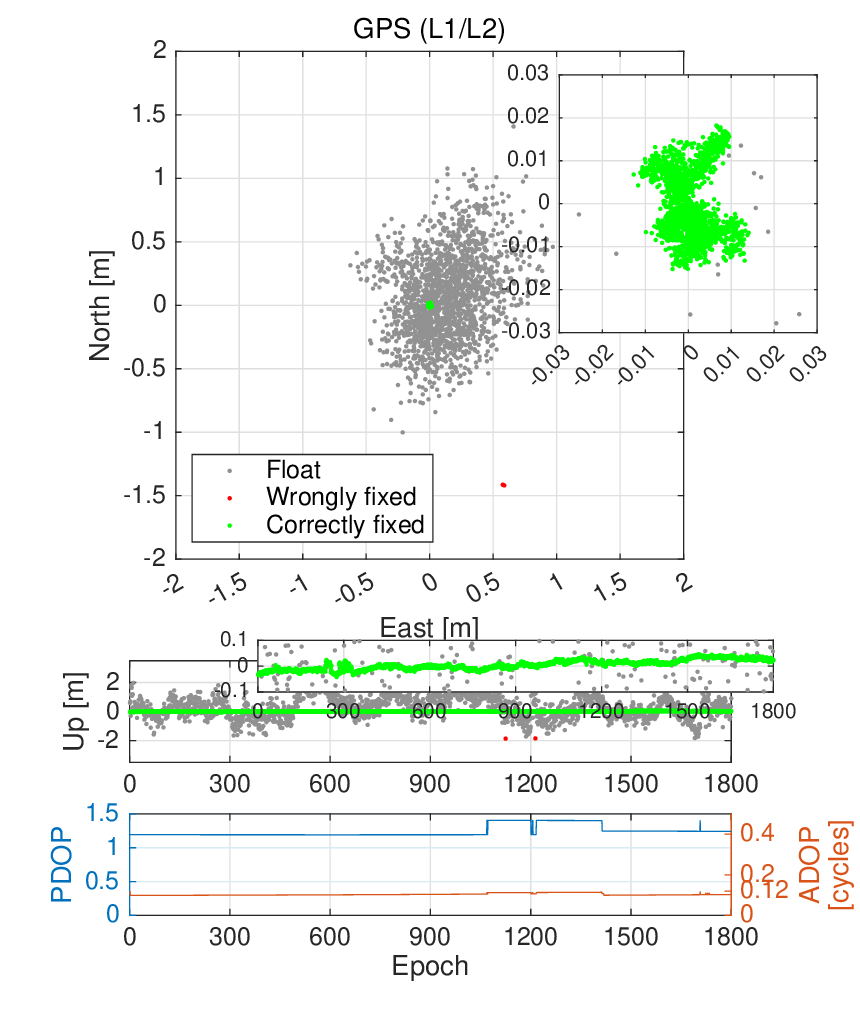}%
    \label{fig:res_gps2}}
    \subfloat[]{\includegraphics[width=.33\linewidth]{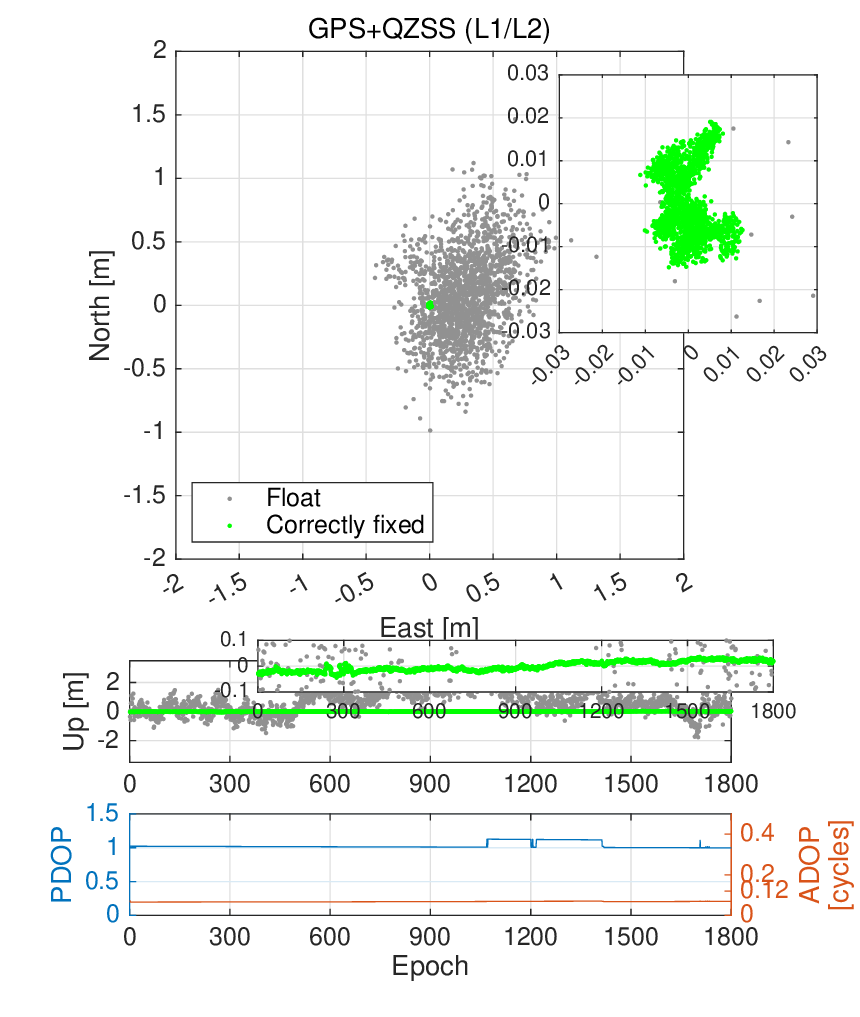}%
    \label{fig:res_dual}}
    \vfill
    \subfloat[]{\includegraphics[width=.33\linewidth]{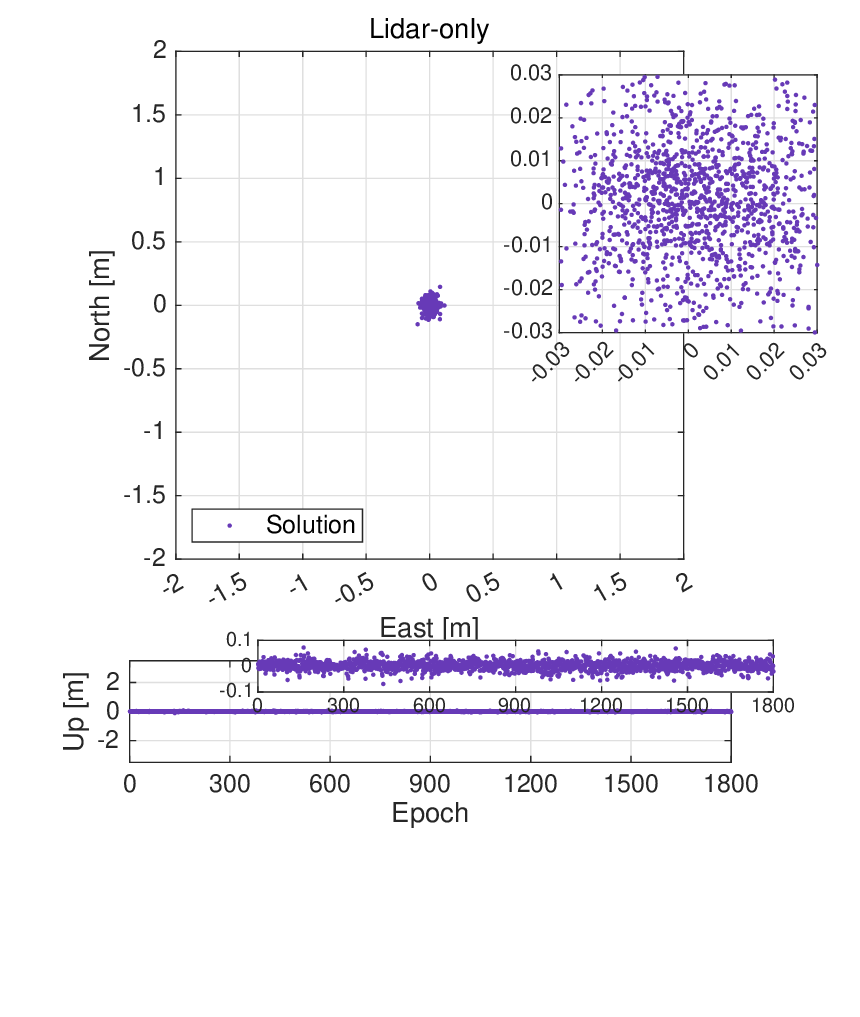}%
    \label{fig:res_lidar}}
    \subfloat[]{\includegraphics[width=.33\linewidth]{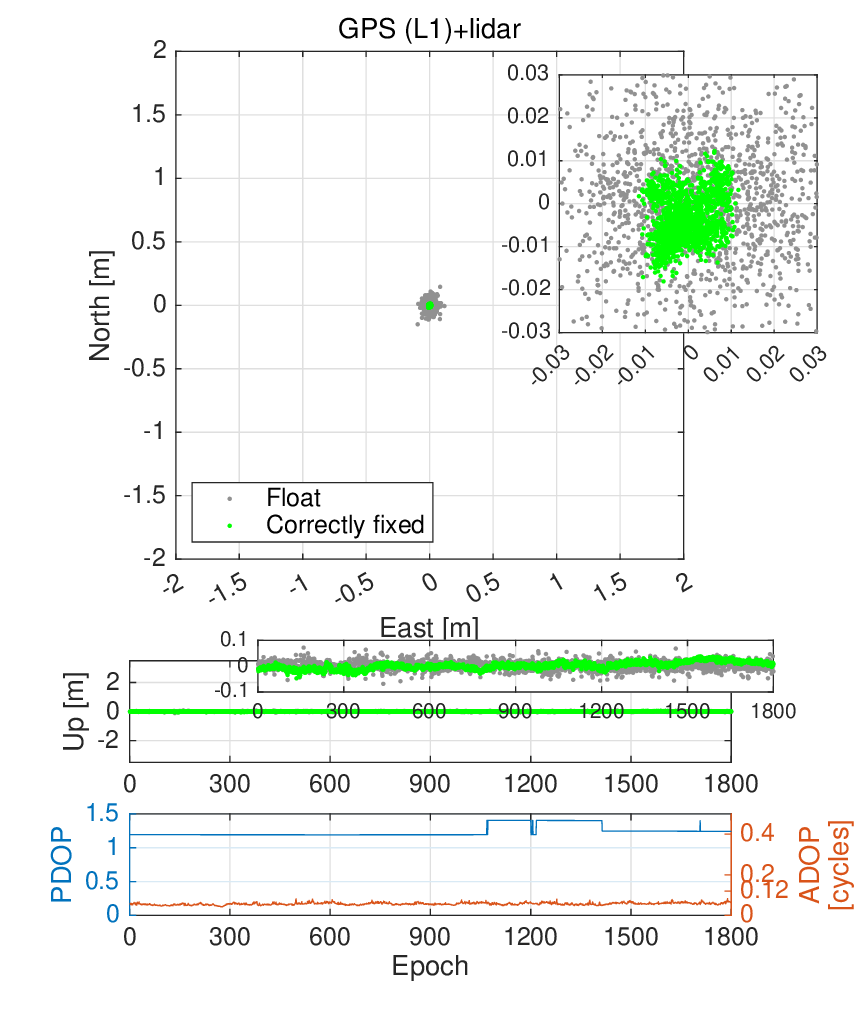}%
    \label{fig:res_gpslidar}}
    \subfloat[]{\includegraphics[width=.33\linewidth]{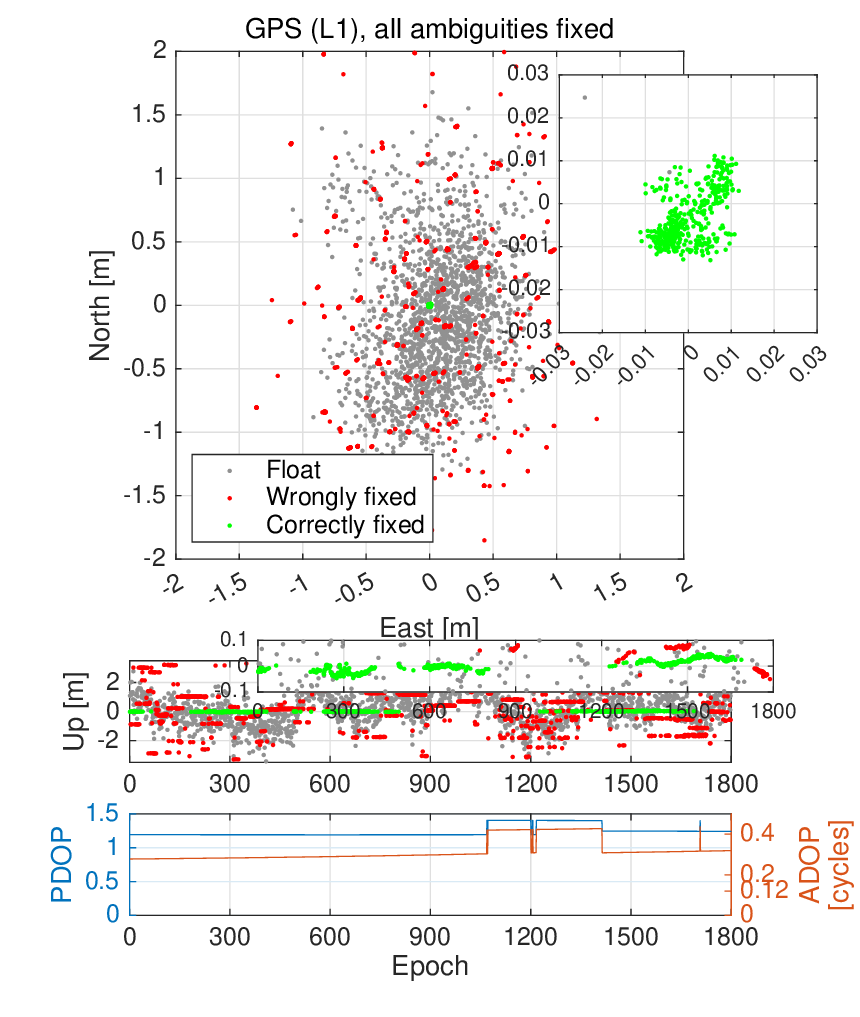}%
    \label{fig:res_gps_far}}
    \caption{ENU errors, PDOP and ADOP of positioning results from all tested methods (ambiguities are fixed when $Ps\geqslant99.9\%$ for \textit{a} to \textit{e}). Grey: float solutions; red: wrongly-fixed solutions; green: correctly-fixed solutions; purple: positioning solutions for Lidar-only method. (a) GPS (L1) results. (b) GPS (L1/L2) results. (c) GPS+QZSS (L1/L2) results. (d) Lidar-only results. (e) GPS (L1)+lidar results. (f) GPS (L1) results by fixing all ambiguities.}
    \label{fig:res}
\end{figure*}

Positioning errors per epoch, time series of the Position Dilution of Precision (PDOP) and ADOP of the tested methods are provided in Fig.~\ref{fig:res}. In addition, Table \ref{tab:posres} illustrates the horizontal, vertical and 3D RMSE of the 5 positioning approaches. It is shown that for the positioning duration, 6 to 7 GPS satellites and 3 QZSS satellites can be tracked. According to the analysis using ADOP in Section \ref{sec:analysis}, at least 8 satellites are required to achieve the success rate of 99.9\% with one frequency. This corresponds with the GPS (L1) results in Fig.~\ref{fig:res_gps}, in which all of the solutions are float because of high ADOP (or low success rates). Fig.~\ref{fig:ADOP_Ps_m_44} also suggested that the minimum number of satellites needed for ADOP $\leqslant$ 0.12 cycles with two frequencies is 5, which agrees with Fig.~\ref{fig:res_gps2}, in which all solutions are fixed for GPS (L1/L2) since sufficient satellites are observed. However, two epochs are found with wrongly fixed solutions when the numbers of satellites decrease from 7 to 6. Unsurprisingly, the accuracy of GPS (L1) is the lowest among all, giving a meter-level 3D RMSE of 1.355 m, whereas GPS (L1/L2) improves it to 0.083 m because of having more observations to enable IAR. In comparison, GPS+QZSS (L1/L2) can successfully fix integer ambiguities for all epochs due to the additional satellites, significantly improving the precision of the positioning results. The horizontal and 3D RMSE also dramatically decrease to 0.009 m and 0.02 m, offering millimeter- to centimeter-level accuracy.

Moving on to the positioning methods with the integration of lidar measurements, although lidar registration is considered highly accurate in this experiment (Fig.~\ref{fig:sre}), the horizontal, vertical and 3D RMSE of the Lidar-only positioning solutions are 0.034 m, 0.017 m and 0.038 m, respectively, which are less accurate than GPS+QZSS (L1/L2). However, as predicted in Fig.~\ref{fig:ADOP_eval_high_f1_n44}, with abundant lidar keypoints with decimeter-level precision and more than two GNSS satellites, single-frequency IAR is feasible. Fig.~\ref{fig:res_gpslidar} shows that by integrating lidar and GPS (L1) observations, ADOP is always below 0.12 cycles and all ambiguities are correctly fixed, giving the positioning accuracy that is even higher than GPS+QZSS (L1/L2) thanks to the contribution of lidar, with the horizontal, vertical and 3D RMSE of 0.008 m, 0.014 m and 0.016 m.

\begin{figure}[!t]
    \centering
    \subfloat[]{\includegraphics[width=.8\linewidth]{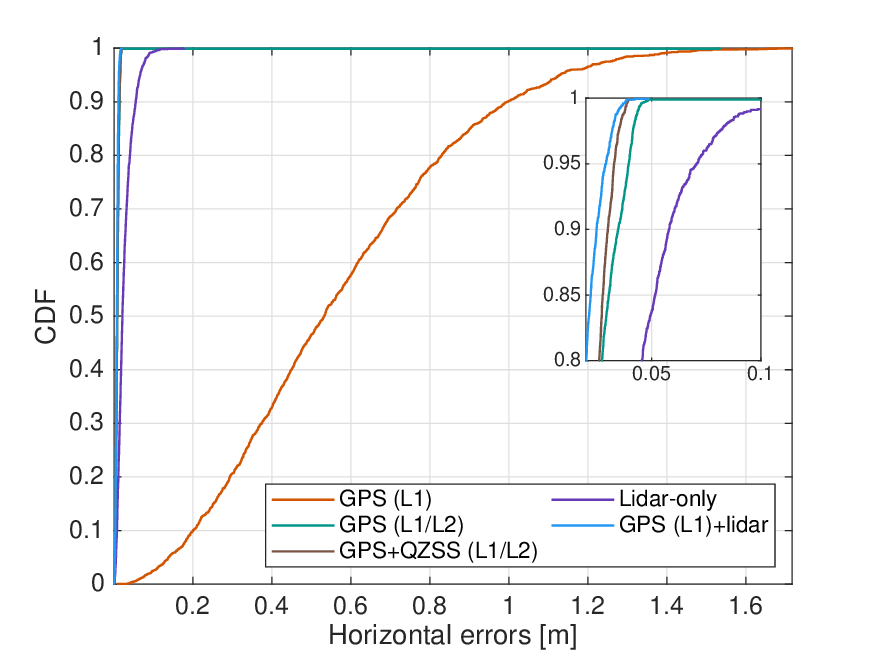}%
    \label{fig:cdf2d}}
    \vfill
    \subfloat[]{\includegraphics[width=.8\linewidth]{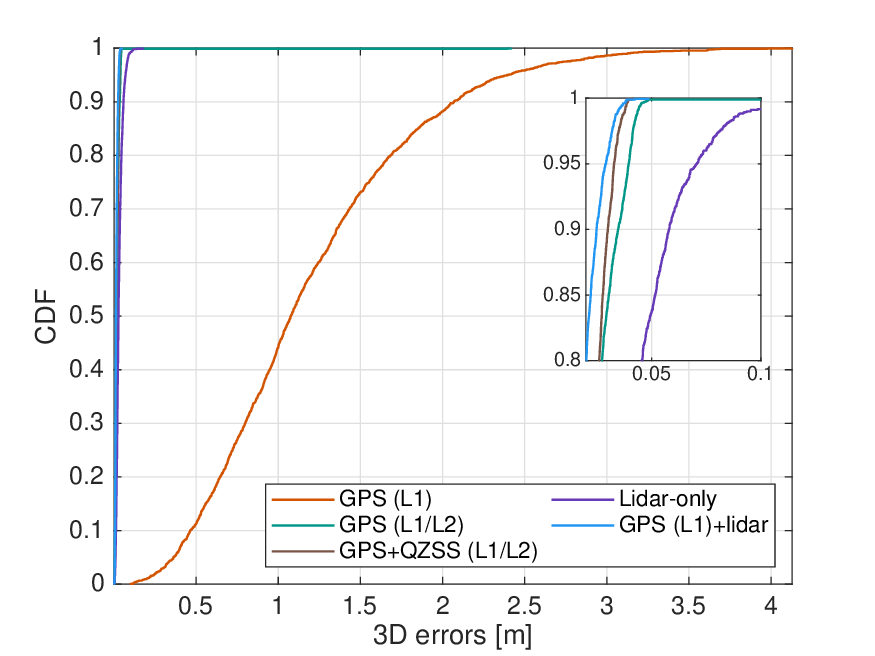}%
    \label{fig:cdf3d}}
    \caption{Cumulative distribution functions of 2D and 3D errors of positioning solutions from all tested methods. (a) 2D CDF. (b) 3D CDF.}
    \label{fig:cdf}
\end{figure}

\begin{figure}[!t]
    \centering
    \includegraphics[width=.8\linewidth]{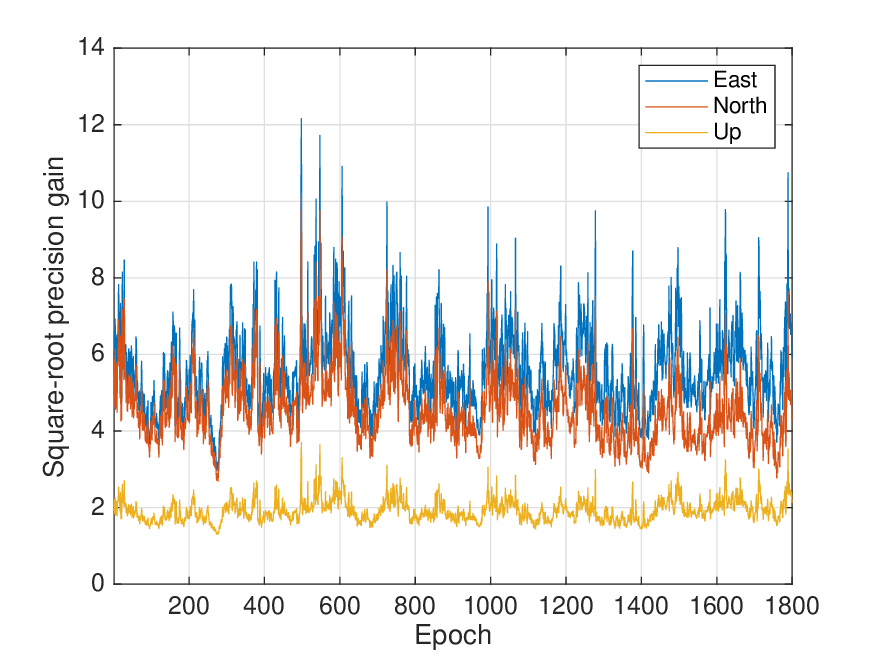}
    \caption{Square-root precision gain of GPS (L1)+lidar positioning solutions (ambiguities are fixed when $Ps\geqslant99.9\%$).}
    \label{fig:gain}
\end{figure}

Fig.~\ref{fig:cdf} presents the cumulative distribution function (CDF) of the horizontal and 3D errors of the tested methods. The superiority of the proposed lidar-aided method in terms of positioning accuracy is clearly shown, outperforming all the other positioning methods. Moreover, to demonstrate the precision improvement of the GPS (L1)+lidar solutions brought by the lidar-aided ambiguity resolution, we evaluate the square-root precision gain in the ENU directions in Fig.~\ref{fig:gain}, which shows how many times the precision of positioning solutions increases by fixing integer ambiguities. Due to the considerably larger number of measurements and higher precision of lidar than the code observations, the float solutions are almost the same as the Lidar-only ones, which can be observed from Fig.~\ref{fig:res_lidar} and \ref{fig:res_gpslidar}. However, by correctly fixing the integer ambiguities, the carrier phase observations further improve the positioning precision by around 6 times horizontally and 2 times vertically, which explains the higher accuracy and precision of GPS (L1)+lidar results than those of Lidar-only as shown in Table \ref{tab:posres} and Fig.~\ref{fig:res_gpslidar}. Note that the precision of Lidar-only solutions in 3 directions are homogeneous in this experiment, and the lower vertical precision gain is caused by the lower precision of GNSS observations in the Up direction.

\subsection{Ambiguity resolution performance}

In order to assess the ambiguity resolution performance in terms of the proportion of correctly fixed epochs, full ambiguity resolution is applied by removing the acceptance test and forcing the resolved integer ambiguities to be fixed for all epochs for the demonstrated positioning methods. Fig.~\ref{fig:res_gps_far} shows positioning errors per epoch, numbers of tracked satellites and ADOP time series of GPS (L1) results, with the empirical success rate computed as 22.7\% and the mean formal success rate evaluated as 48.2\%. Note that GPS+QZSS (L1/L2) and Lidar-only results are not applicable for empirical success rate evaluation since the former provides the benchmark ambiguities and the latter does not utilize GNSS observations. The results of GPS (L1/L2) and the proposed method, namely GPS (L1)+lidar, are also omitted as they are identical to those in Fig.~\ref{fig:res_dual}~and~\ref{fig:res_gpslidar} since all fixed solutions are already accepted when the acceptance test is present. The empirical success rates of these two methods are 99.9\% and 100\%, respectively, while their formal success rates are both computed as above 99.9\%. Again, we have previously concluded that IAR with single-frequency GNSS-only observations from fewer than 8 satellites is not feasible, which is reflected here by the low empirical success rate of GPS (L1). In comparison, the integration of lidar data substantially increases the empirical success rate without requiring additional GNSS measurements and correctly fixes all the integer ambiguities, while keeping ADOP below 0.12 cycles and achieving comparable ambiguity resolution performance as the GNSS-only model using dual-frequency data. In addition, in Section \ref{sec:ADOPeval} it was found that the proposed method can enable IAR using only 2 or 3 satellites. By limiting the number of satellites to 2, which renders GNSS-only methods impossible, the GPS (L1)+lidar method achieves an empirical success rate of 96.8\%, as well as the 2D and 3D RMSE of 0.026 m and 0.033 m, respectively, confirming the theoretical expectation.
\section{Discussion \label{sec:discussion}}

\subsection{Quality of lidar measurements and HD map}

Although we have shown analytically that decimeter-level precision of lidar measurements can enable successful IAR (Fig.~\ref{fig:ADOP_eval_high}), one limitation of our experiment is that we have used highly accurate lidar data, as the SRE of the registered keypoints suggested (Fig.~\ref{fig:sre}). In terms of the HD map, the reference scans are georeferenced with the validated ground truth information provided in the KITTI dataset. In practice, for large-scale HD map products of urban road environments, such accurate georeferencing can be difficult to achieve and they may be produced with larger errors, decreasing the accuracy of the derived lidar measurements. On the other hand, since the reference scans are meant to be acquired from a previous time, the registration accuracy may be influenced by the environmental differences between the rover and reference scans if the HD map is not up-to-date, which is not reflected in our experiment. Furthermore, due to the lack of GNSS raw observations synchronized with the lidar data in KITTI, the rover-reference scan pairs are pre-assigned with a 3 s interval, which is equivalent to the distance of a few meters. In reality, the nearest reference scan from the HD map should be identified using position estimates of the vehicle in real time from less demanding techniques such as SPP.

\subsection{Number of lidar keypoints and empirical success rate}

\begin{figure}[!t]
    \centering
    \includegraphics[width=.8\linewidth]{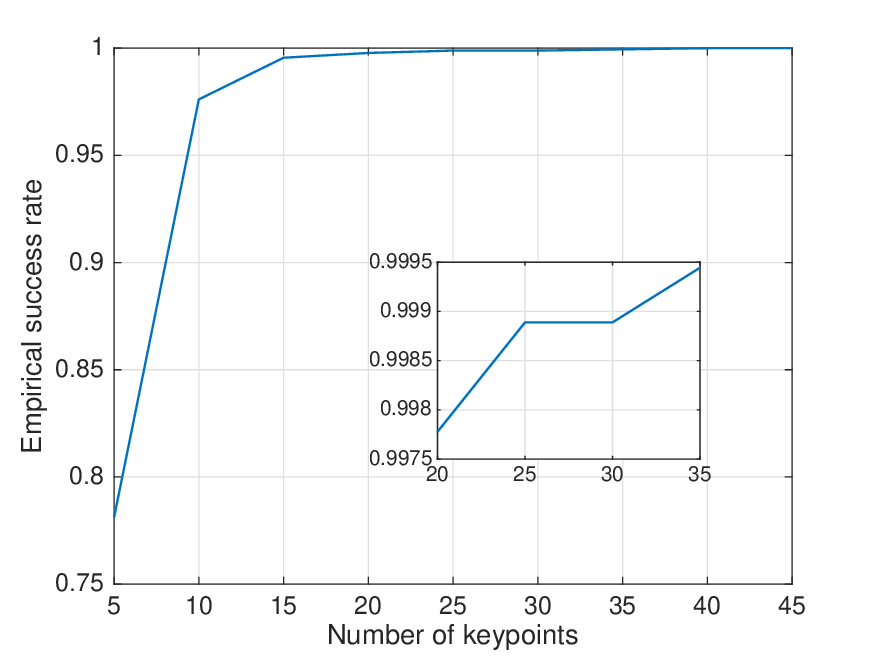}
    \caption{Empirical success rates using different numbers of keypoints and full ambiguity resolution.}
    \label{fig:esr_n}
\end{figure}

It has been demonstrated in Section~\ref{sec:analysis} that the number of correctly matched keypoints is not required to be large to ensure the ambiguity success rate upper bound of 99.9\%, provided that the precision of the lidar measurements is at the decimeter-level, or that a few satellites can be tracked. In order to determine a recommended number of lidar keypoints for consistently successful IAR in a practical environment, we have repeated the experiment using the GPS (L1)+lidar positioning setup with the number of lidar keypoints limited to a range between 5 and 45 by applying full ambiguity resolution. The empirical success rates against different numbers of keypoints are shown in Fig.~\ref{fig:esr_n}, which indicates that the empirical success rate is beyond 99\% with only 10 keypoints, and increases to above 99.9\% when 35 or more keypoints are used. It should be remarked that this quantity of correspondences between registered rover and reference scans can be easily obtained, as the empirical minimum number of keypoints per epoch in our experiment is 44.

\subsection{Runtime efficiency}

The keypoint matching and positioning stages of the proposed method are experimented with the Torch-Points3D \cite{chaton2020torch} implementation of Ms-SVConv and MATLAB \cite{MATLAB}, respectively. On a platform consisting of AMD Ryzen 3800XT CPU and NVIDIA RTX 3070 GPU, the former approximately takes 0.85 s and the latter takes 0.05 s to complete the computation for each epoch. Therefore, the proposed instantaneous lidar-aided ambiguity resolution method has the potential of near-real-time positioning for vehicles.

\section{Concluding remarks \label{sec:concl}}

In this contribution we proposed an instantaneous lidar-aided ambiguity resolution method focusing on vehicle positioning in urban canyons, where GNSS signals are prone to blockage and multipath. The lidar measurements are generated by a keypoint extraction strategy via learning-based point cloud registration between rover scans and reference scans from a pre-built HD map. A mixed measurement model is employed to integrate such lidar measurements with their DD GNSS counterparts at the observation-level to obtain precise float solutions and enable instantaneous IAR. Closed-form expressions of the ambiguity variance matrix and the corresponding ADOP are developed to provide a priori evaluation of the ambiguity resolution performance using the numbers of available satellites and keypoints, as well as the precision of the measurements (cf. \ref{eq:ADOPratio}).

Our analytical study has shown when limited GNSS satellites and/or frequencies are accessible, which is often the case in urban environments, the proposed lidar-aided method can significantly reduce the ADOP value comparing with the GNSS-only approach, thus enabling successful instantaneous IAR (Fig.~\ref{fig:ratio}). Moreover, a moderate number of lidar keypoints can reduce the required satellites to track, to the extent that IAR is feasible with single-frequency data from only 2 or 3 satellites (Fig.~\ref{fig:ADOP_eval_high}). The numerical results from a simulated experiment illustrate that the proposed method achieves the empirical ambiguity success rate of 100\% and 3D positioning RMSE of 0.017 m using GPS L1 and lidar measurements, thereby outperforming both the Lidar-only and GPS+QZSS (L1/L2) positioning methods (Fig.~\ref{fig:res}). Future work will undertake real-world experiments to further examine the performance of the proposed method in GNSS-challenging urban canyons.


\section*{Acknowledgments}
This research did not receive any specific grant from funding agencies in the public, commercial, or not-for-profit sectors. The first author acknowledges the financial support from The University of Melbourne through the Melbourne Research Scholarship.

\section*{Appendix: Supplementary proofs\label{sec:appendix}}
An application of the determinant factorization rule~\cite{odijk2008adop,koch1999} to the ambiguity variance matrix (\ref{eq:Qa}) gives
\be
\label{eq:app1}
\ba{lcl}
|\bm{Q_{\hat{a}\hat{a}}}|  &=&  \dfrac{(\sqrt{2}\,\sigma_{\phi})^{2f(m-1)}}{|\bm{\Lambda}|^{2(m-1)}}\,|\bm{D}^{\top}\bm{W}_G^{-1}\bm{D}|^{f}\vspace{1mm}\\
& & \times\; |\bm{I}_3+\frac{\sigma_{p}^2}{\sigma_{\phi}^2}\bm{Q}_{\bm{\hat{b}\hat{b}},G}^{-1}\bm{Q_{\hat{b}\hat{b}}}|
\ea
\ee
Substitution of the identities $|\bm{\Lambda}|=\bar{\lambda}^f$, $|\bm{D}^{\top}\bm{W}_G^{-1}\bm{D}|=w_o^{2(m-1)}$~\cite{odijk2008adop}, and the phase-to-code variance ratio $\epsilon= (\sigma_{\phi}/\sigma_{p})^2$ into (\ref{eq:app1}) yields 
\be
\label{eq:app2}
\ba{lcl}
|\bm{Q_{\hat{a}\hat{a}}}|  &=&  \left[\dfrac{\sqrt{2}\, w_o\,\sigma_{\phi}}{\bar{\lambda}}\right]^{2f(m-1)}\,\vspace{1mm}\\
& & \times\; |\bm{I}_3+\frac{1}{\epsilon}\bm{Q}_{\bm{\hat{b}\hat{b}},G}^{-1}\bm{Q_{\hat{b}\hat{b}}}|
\ea
\ee
The above expression for the determinant of the ambiguity variance matrix, together with the ADOP definition (\ref{eq:adop}), gives
\be
\label{eq:appADOP}
{\rm ADOP} = \sqrt{2}w_o \dfrac{\sigma_\phi}{\bar{\lambda}}\;
|\bm{I}_3+\frac{1}{\epsilon}\bm{Q}_{\bm{\hat{b}\hat{b}},G}^{-1}\bm{Q_{\hat{b}\hat{b}}}|^{\frac{1}{2f(m-1)}}
\ee
\noindent For the GNSS-only case, we have $\bm{Q_{\hat{b}\hat{b}}}=\bm{Q}_{\bm{\hat{b}\hat{b}},G}$. This simplifies the last term in (\ref{eq:appADOP}) to $|\bm{I}_3+\frac{1}{\epsilon}\bm{Q}_{\bm{\hat{b}\hat{b}},G}^{-1}\bm{Q_{\hat{b}\hat{b}}}|^{\frac{1}{2f(m-1)}}=[1+\frac{1}{\epsilon}]^{\frac{3}{2f(m-1)}}$, from which the ADOP of the GNSS-only model (\ref{eq:ADOPg}) follows. For the integrated GNSS-lidar case however, we have $\bm{Q_{\hat{b}\hat{b}}}^{-1}=\bm{Q}_{\bm{\hat{b}\hat{b}},G}^{-1}+\bm{Q}_{\bm{\hat{b}\hat{b}},L}^{-1}$ instead (cf.~\ref{eq:fQ}). The first expression of the ADOP-ratio (\ref{eq:ADOPratio}) follows then by substituting $\bm{Q}_{\bm{\hat{b}\hat{b}},G}^{-1}=\bm{Q}_{\bm{\hat{b}\hat{b}}}^{-1}-\bm{Q}_{\bm{\hat{b}\hat{b}},L}^{-1}$ into $|\bm{I}_3+\frac{1}{\epsilon}\bm{Q}_{\bm{\hat{b}\hat{b}},G}^{-1}\bm{Q_{\hat{b}\hat{b}}}|^{\frac{1}{2f(m-1)}}$, showing that
\be
\ba{l}
|\bm{I}_3+\frac{1}{\epsilon}\bm{Q}_{\bm{\hat{b}\hat{b}},G}^{-1}\bm{Q_{\hat{b}\hat{b}}}|^{\frac{1}{2f(m-1)}}=[1+\frac{1}{\epsilon}]^{\frac{3}{2f(m-1)}}\vspace{1mm}\\
\qquad \times\; |\bm{I}_3-\dfrac{1}{1+\epsilon}\bm{Q}_{\bm{\hat{b}\hat{b}},L}^{-1}\bm{Q_{\hat{b}\hat{b}}}|^{\frac{1}{2f(m-1)}}
\ea
\ee
Finally, the second expression of the ADOP-ratio (\ref{eq:ADOPratio}) follows from the definition of the generalized eigenvalues (\ref{eq:eigenvalues}), that is
\be
|\bm{I}_3-\dfrac{1}{1+\epsilon}\bm{Q}_{\bm{\hat{b}\hat{b}},L}^{-1}\bm{Q_{\hat{b}\hat{b}}}| = \prod\limits_{i=1}^3\left[1-\frac{1}{1+\epsilon}\frac{1}{\gamma_i}
    \right]
\ee

\bibliographystyle{IEEEtran}
\bibliography{references}

\vfill

\end{document}